\documentclass[10pt,twocolumn,letterpaper]{article}
\usepackage{iccv}              

\definecolor{iccvblue}{rgb}{0.21,0.49,0.74}
\usepackage[breaklinks,colorlinks,allcolors=iccvblue]{hyperref}

\def\paperID{3420} 
\def\confName{ICCV}
\def\confYear{2025}

\title{
\includegraphics[scale=0.2]{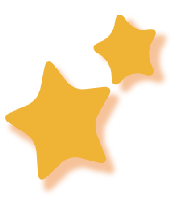} Pose-Star: Anatomy-Aware Editing for Open-World Fashion Images}

\author{
  Yuran Dong, Mang Ye\thanks{Corresponding Author: Mang Ye (yemang@whu.edu.cn)} \\  
   National Engineering Research Center for Multimedia Software \\                    
  School of Computer Science, Wuhan University, Wuhan, China \\            
  {\tt\small \{dongyuran, yemang\}@whu.edu.cn}  \\
  {\small{\url{https://github.com/NDYBSNDY/Pose-Star}}}
}

\begin{document}

\maketitle
\begin{abstract}
To advance real-world fashion image editing, we analyze existing two-stage pipelines—mask generation followed by diffusion-based editing—which overly prioritize generator optimization while neglecting mask controllability. This results in two critical limitations: I) poor user-defined flexibility (coarse-grained human masks restrict edits to predefined regions like upper torso; fine-grained clothes masks preserve poses but forbid style/length customization). II) weak pose robustness (mask generators fail due to articulated poses and miss rare regions like waist, while human parsers remain limited by predefined categories).
To address these gaps, we propose \textbf{Pose-Star}, a framework that dynamically recomposes body structures (e.g., neck, chest, etc.) into anatomy-aware masks (e.g., chest-length) for user-defined edits. 
In Pose-Star, we calibrate diffusion-derived attention (Star tokens) via skeletal keypoints to enhance rare structure localization in complex poses, suppress noise through phase-aware analysis of attention dynamics (Convergence→Stabilization→Divergence) with threshold masking and sliding-window fusion, and refine edges via cross-self attention merging and Canny alignment. 
This work bridges controlled benchmarks and open-world demands, pioneering anatomy-aware, pose-robust editing and laying the foundation for industrial fashion image editing.

\end{abstract}    
\section{Introduction}
\label{sec:intro}

Fashion image editing~\cite{Zeng2022FlowFaceSF, Chen2024ZeroshotIE} enables critical applications like virtual try-on~\cite{Choi2024ImprovingDM, Chong2024CatVTONCI} and clothes design~\cite{jiang2025vidsketch, yang2025videograin}, yet existing methods—trained on curated datasets~\cite{Morelli2022DressCH, Xiao2017FashionMNISTAN, Rostamzadeh2018FashionGenTG,
Huang2023FIRSTAM}—prioritize generation fidelity over real-world adaptability. So, open-world fashion editing faces challenges:  user-defined region flexibility (\emph{e.g., customizing clothes lengths}), pose robustness (\emph{e.g., extreme articulations}), in-the-wild generalizability (\emph{e.g., cluttered scenes}).

\begin{figure}[t]
  \centering
   \includegraphics[width=1\linewidth]{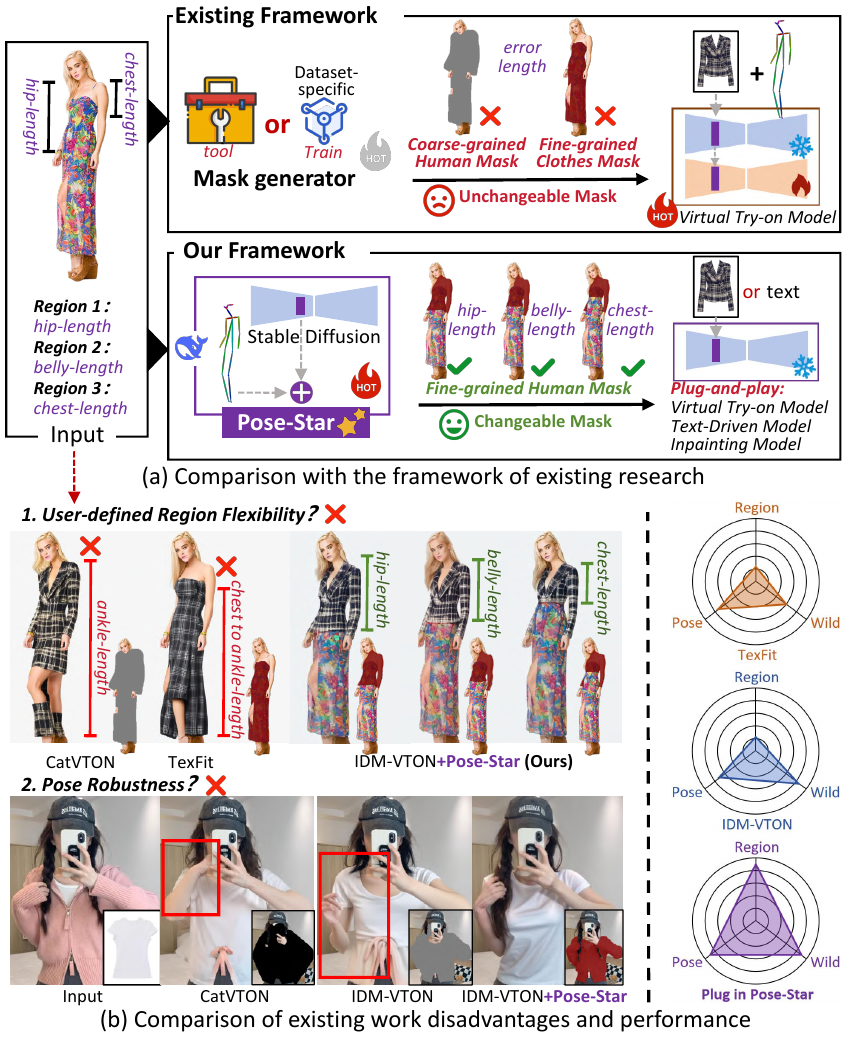}
   \caption{
Problem \sethlcolor{red1}\hl{\textbf{\emph{I)}}}: Existing methods lack anatomy-aware control, limiting user-defined regions.
Problem \emph{\sethlcolor{red1}\hl{\textbf{II)}}}: Articulated poses induce segmentation errors, distorting edits and robustness.
Evaluation based on real-world challenges: \textbf{Region:} User-defined Region Flexibility, \textbf{Pose:} Pose Robustness, and \textbf{Wild:} In-the-Wild Generalizability. 
Refer to Sec.\ref{Effectiveness} and \cref{fig:table} for details.
}
   \label{fig:intro}
\end{figure}

\begin{figure*}[t]
  \centering
    \captionsetup{type=figure}
    \includegraphics[width=1\textwidth]{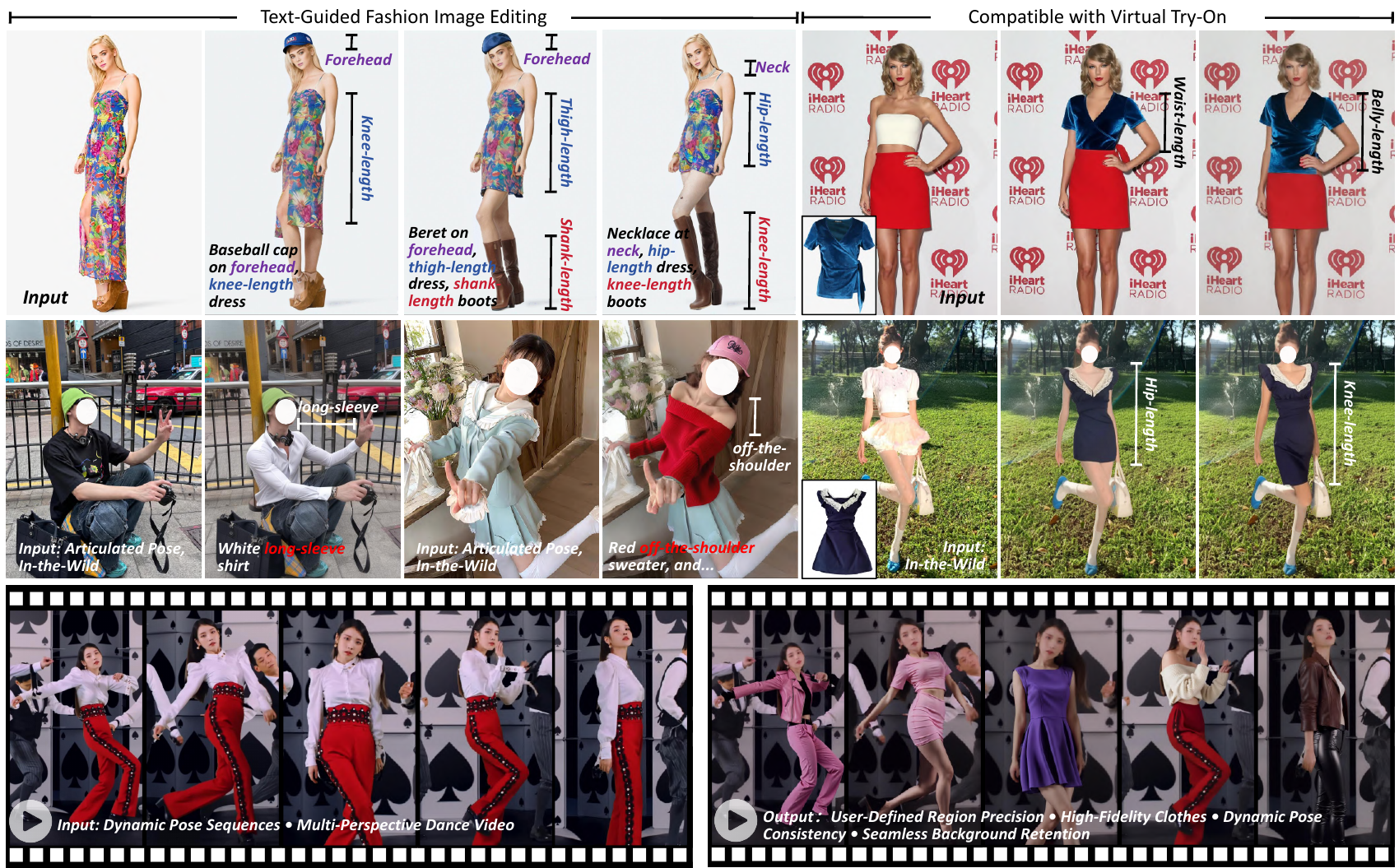}
    \captionof{figure}{
    Pose-Star is compatible with text-guided fashion editing and visual try-on, enabling unprecedented capabilities: customizable clothes length adjustments, precise edits on highly articulated poses, and robust adaptation to wild scenarios with cluttered backgrounds. Notably, it extends to video editing applications, seamlessly modifying dance sequences with complex poses and dynamic camera angles.
    }
    \label{fig:one}
\end{figure*}

Current fashion-specific frameworks~\cite{Baldrati2023MultimodalGD, Gou2023TamingTP, Lee2022HighResolutionVT,
Lin2023FashionTexCV} adopt a two-stage pipeline (\cref{fig:intro}a): mask-guided region specification followed by diffusion-based editing. The mask generator types include:
Coarse-grained human parser masks~\cite{Choi2024ImprovingDM,Xu2024OOTDiffusionOF}, restrict edits to predefined zones (\emph{e.g., upper torso}), preventing length customization or partial clothes retention (\cref{fig:intro}b).
Fine-grained clothes masks (train-based)~\cite{wang2024dpdedit, Pernu2023FICETF, Wang2024TexFitTF} rigidly adhere to clothes boundaries, making style changes (\emph{"dress → pants"}) impossible.
These methods relying on open tools or fashion datasets fail in handling user-defined edits due to inflexible anatomical control and limited in-the-wild generalizability. This raises a pivotal problem: \emph{\sethlcolor{red1}\hl{\textbf{I)} how to design a dynamic, anatomy-aware mask generator tailored for open-world fashion editing to achieve user-defined region flexibility?}}

Initial attempts to generate anatomy-aware masks (\emph{e.g., belly-length}) via open-world segmenters~\cite{Qin2022HighlyAD,Qin2020U2NetGD} failed due to their foreground-background paradigm lacking anatomical granularity. Open-world detectors~\cite{Jiang2024TRex2TG} localized coarse structures (\emph{e.g., neck}) but collapsed on rare regions (\emph{e.g., waist}) and articulated poses due to noun-centric training. Human parsers~\cite{Yang2023DeepLT,Zhang2022AIParsingAI} further exhibited limited anatomical coverage (\emph{e.g., missing chest}).
These failures reveal a critical gap: existing methods cannot robustly segment anatomical structures or detect rare regions under complex poses due to inherent knowledge limitations. This raises a pivotal problem: \emph{\sethlcolor{red1}\hl{\textbf{II)} how to precisely localize fine-grained body structures and aggregate noise-free regions in open-world scenarios to achieve pose robustness for detection?}}

To address the above problems, this work pioneers the study of open-world fashion-specific mask generation. For problem \sethlcolor{red1}\hl{\textbf{\emph{I)}}}, we propose \textbf{Pose-Star}, a \textbf{plug-and-play}, \textbf{training-free}  framework that synergizes the style adaptability of coarse-grained human masks with the pose precision of fine-grained clothes masks by dynamically generating fine-grained human masks guided by user-defined instructions (\emph{e.g., "belly-length"}). The core innovation lies in decomposing anatomical prompts into local body structures (\emph{e.g., "belly-length"→ neck, chest, waist, etc.}) and executing localization, aggregation, and optimization to obtain the target human mask. For problem \sethlcolor{red1}\hl{\textbf{\emph{II)}}}, Pose-Star implementation involves: 1) Localization: Pose-guided calibration of diffusion-derived attention tokens (Star tokens) via skeletal keypoints enhances precision for rare anatomical regions (\emph{e.g., waist}) in complex poses; 2) Aggregation: We analysis of diffusion attention dynamics (convergence→stabilization→divergence), based on which we design threshold masking and sliding windows to preserve phase-stability regions and eliminate noise; 3) Optimization: Design of cross-self attention merge and Canny alignment to calibrate mask boundaries based on our observation of the complementary roles of cross-attention (spatial focus) and self-attention (edge coherence).
Pose-Star achieves unprecedented functionality; selected example in \cref{fig:one}.
Our principal contributions are summarized as:

\begin{itemize}
    \item[ ] \includegraphics[scale=0.07]{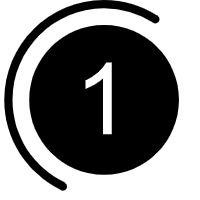} 
    \textbf{\emph{Plug-and-Play Framework: Pose-Star.}} A mask generator that bridges the critical gap in user-defined region flexibility by dynamically synthesizing anatomy-aware masks, paving the way for industrial applications of fashion image editing.
    \item[ ] \includegraphics[scale=0.07]{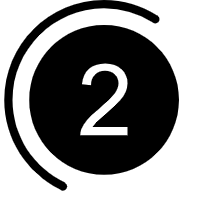} \textbf{\emph{Technical Highlights.}} We synergize pose-guided calibration of diffusion tokens (via keypoints) for rare anatomical localization, phase-aware aggregation (threshold masking and sliding windows) to suppress noise across convergence/divergence phases, and cross-self attention merging with Canny alignment to refine edges, achieving pose robustness for anatomy-aware masks.    
    \item[ ] \includegraphics[scale=0.07]{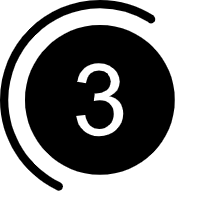} 
    \textbf{\emph{Real Open-World Usability Validation.}} Demonstrating state-of-the-art performance in user-defined flexibility, pose robustness, and wild generalizability.
\end{itemize}

These contributions establish a novel paradigm for anatomy-aware fashion editing, transitioning the field from dataset-bound optimization to open-world controllability.
Refer to Appendix \ref{Related Work} for related work.

\section{Proposed Method}

To address the inability of existing methods to interpret user-defined anatomical regions (\emph{e.g., "belly-length"}) and handle intricate poses, we propose Pose-Star. As shown in \cref{fig:intro}a and \cref{fig:framework2}, our framework leverages DeepSeek~\cite{DeepSeekAI2024DeepSeekV3TR} (Details in Appendix \ref{DeepSeek}) to parse anatomical specifications into the body structure group, then extracts and fuses body structure regions via the Pose-Star module to generate a Fine-grained Human Mask. This mask guides pose-robust editing in plug-and-play editors (\emph{e.g.,} visual try-on model IDM-VTON~\cite{Choi2024ImprovingDM}, inpainting model PowerPaint~\cite{Zhuang2023ATI}), ensuring anatomical alignment with user intent. Key challenges: 
\begin{itemize}
    \item  \textbf{C1:} Fine-Grained \textbf{Localization}.
How to precisely localize anatomized structures (\emph{e.g., "belly-length" includes neck, chest, waist, etc.}) in images with complex poses?
    \item \textbf{C2:} Noise-Free Structures \textbf{Aggregation}.
How to eliminate noise while aggregating multiple body structures regions to retain only semantically consistent areas?
    \item \textbf{C3:} Contour Precision \textbf{Optimization}.
How to refine the edge accuracy of the generated human mask to align with anatomical boundaries?
\end{itemize}
For each challenge, we propose solutions corresponding to Sec.\ref{C1-Location}, \ref{C2-Region}, and \ref{C3-Contour}, respectively.

\begin{figure*}[t]
  \centering
   \includegraphics[width=0.95\linewidth]{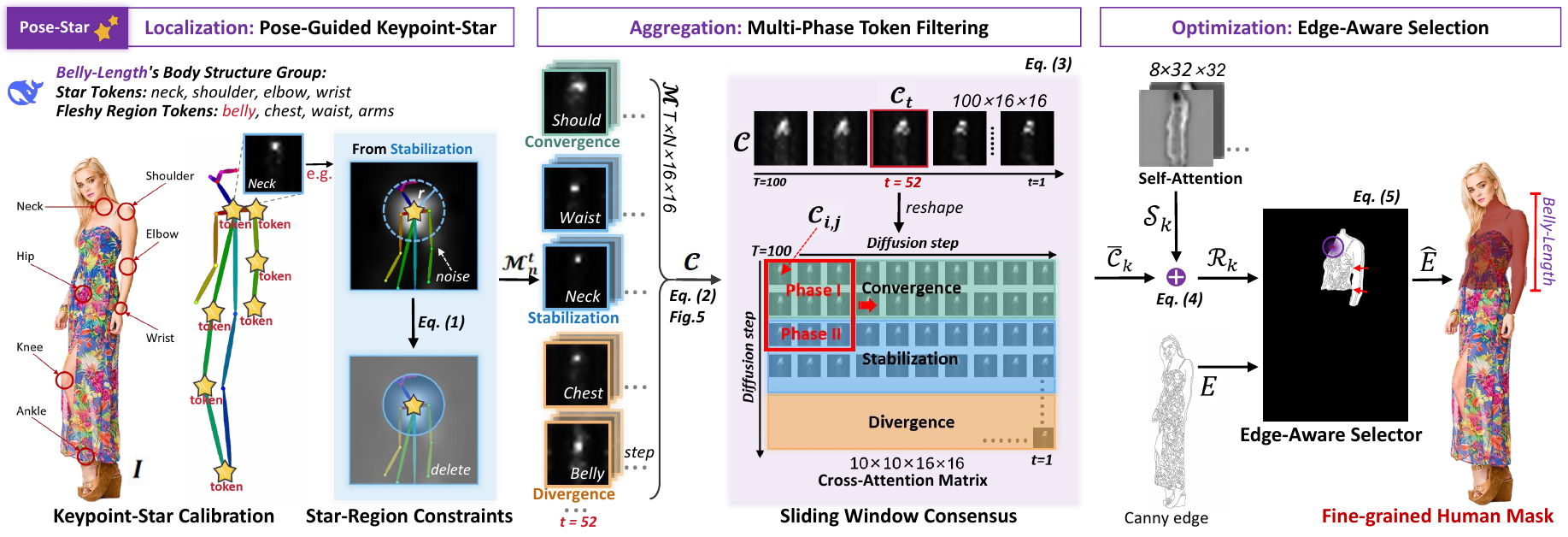}
   \caption{Pose-Star Framework. The generation of the \emph{Bell-Length} mask as an example. Best viewed in color. Zoom in for details.}
   \label{fig:framework2}
\end{figure*}

\subsection{Localization: Pose-Guided Keypoint-Star}\label{C1-Location}
Initial attempts to localize fine-grained body parts (\emph{e.g., belly, waist}) in complex poses using open-word detectors (YOLO-World~\cite{Cheng2024YOLOWorldRO}, DINO1.5/1.6~\cite{Ren2024GroundingD1}, T-Rex2~\cite{Jiang2024TRex2TG}, GLEE~\cite{Wu2023GeneralOF}) revealed critical limitations: noun-centric training constraints lead to vocabulary rigidity  (\emph{e.g.,} the common words \emph{"hand"} or \emph{"head"} can be detected, but the fine-grained words \emph{"abdomen"} or \emph{"waist"} cannot be distinguished. ), causing failures under articulation. In contrast, U-Net-based Stable Diffusion models~\cite{Rombach2021HighResolutionIS}—trained on 200M+ samples with full-text embeddings—exhibit superior spatial-textual alignment, enabling open-vocabulary localization of anatomical geometry. Leveraging this insight, we propose Pose-Guided Keypoint-Star, which utilizes the semantic comprehension capabilities of diffusion to generate anatomy-aware masks. For example, belly-length localization via attention token calibration (\cref{fig:framework2}):

\textbf{Diffusion-Based Token Initialization.}
For a real image $I$, we first perform DDIM inversion~\cite{Song2020DenoisingDI} followed by Null-text optimization~\cite{Null-text} reconstruction. During reconstruction, consistent with \cite{Hertz2022PrompttoPromptIE}, the Star Token embeddings (\emph{e.g., belly}) are projected as keys/values ($K/V$) in the U-Net's cross-attention layers, while the image latent variables (corresponding to $I$) form queries ($Q$). The $Q-K$ interaction generates an attention map set $M$, where each map corresponds to a Star Token representing a specific anatomical structure, with spatial attention scores indicating target region saliency. This set $M$ serves to initialize the attention maps for Star Tokens. Method details in Appendix \ref{Attention}.

\textbf{Keypoint-Star Calibration.}
While token maps effectively localize anatomical structures in standard poses, their attention patterns degrade under extreme articulation due to skeletal ambiguities. To address this, we introduce Star Tokens derived from OpenPose~\cite{Cao2018OpenPoseRM} skeletal keypoints. 
Specifically, we divide token maps of anatomical regions into Star Token and Fine-grained Token. Star Token corresponds to pose keypoints (\emph{e.g., hip or shoulder joints}). Fine-grained Token represents fine-grained body composition (\emph{e.g., belly or waist}).
The centroid of each star token map is calibrated to match the spatial coordinates of the corresponding keypoint, ensuring robust localization even in complex poses.
Fine-grained token maps are aligned with biomechanically plausible positions through the frame of the pose. 
Clothes tokens are also used for regional references.

\textbf{Star-Region Constraints.}
The refinement process begins by normalizing each star token map to encode pixel-wise anatomical relevance, where values range from 0 (irrelevant regions) to 1 (target regions). Subsequently, a radial constraint is applied: For the center point $(m,n)$ from each star, probabilities outside a biomechanically plausible radius $r$ are suppressed, effectively eliminating noise from occluded limbs or extreme articulations. For each pixel $p_{i,j}$ of the token map, the filtering equation is as follows:
\begin{equation}
\hat{p}_{i,j} = \left\{\begin{matrix}
p_{i,j},  & if \; \sqrt{(i-m)^{2} + (j-n)^{2}} \le r,\\
0,  & otherwise.
\end{matrix}\right.
  \label{eq:important}
\end{equation}

The filtered token tensor $\mathcal{M} \in \mathbb{R} ^{T\times N\times 16\times 16} $ at pixel-level achieves precise localization of fine-grained body structure. Total number of diffusion steps $T = 100$, each step $t$ produces a sub-tensor $\mathcal{M}_{t}  \in \mathbb{R} ^{N\times 16\times 16}$, where $N$ equals the total number of anatomical structures ($N = \;$star tokens + fine-grained tokens). Each token map $\mathcal{M}_{t}^{n}   \in \mathbb{R} ^{16\times 16}$ (for $n\in \left \{ 1,...,N \right \} $) encodes the spatial saliency of a specific body part at diffusion step $t$, ensuring temporally consistent and anatomically grounded localization.

\subsection{Aggregation: Multi-Phase Token Filtering}\label{C2-Region}
The synthesis of user-specified anatomical regions (\emph{e.g., "belly-length"}) requires aggregating fine-grained token maps while suppressing noise artifacts. However, temporal inconsistency in attention maps across diffusion steps poses a critical challenge: the spatial extent of token map attention dynamically evolves during the diffusion process, making it infeasible to select a \emph{fixed step} $t$ that universally aligns with target boundaries of all anatomical structures. 
To study diffuse attention dynamics, we used Null-text~\cite{Null-text} to invert complex poses (from DeepFashion-MultiModal~\cite{jiang2022text2human} or in-the-wild), tracking cross-attention tokens across $T = 100$ steps. Six structures (\emph{e.g., chest, neck, etc.}) were randomly selected for each sample, and labeling first/last target-aligned steps for each structure as phase boundaries, we exposed three-phase evolution (\cref{fig:eq3}): 
\begin{figure}[t]
  \centering
   \includegraphics[width=1\linewidth]{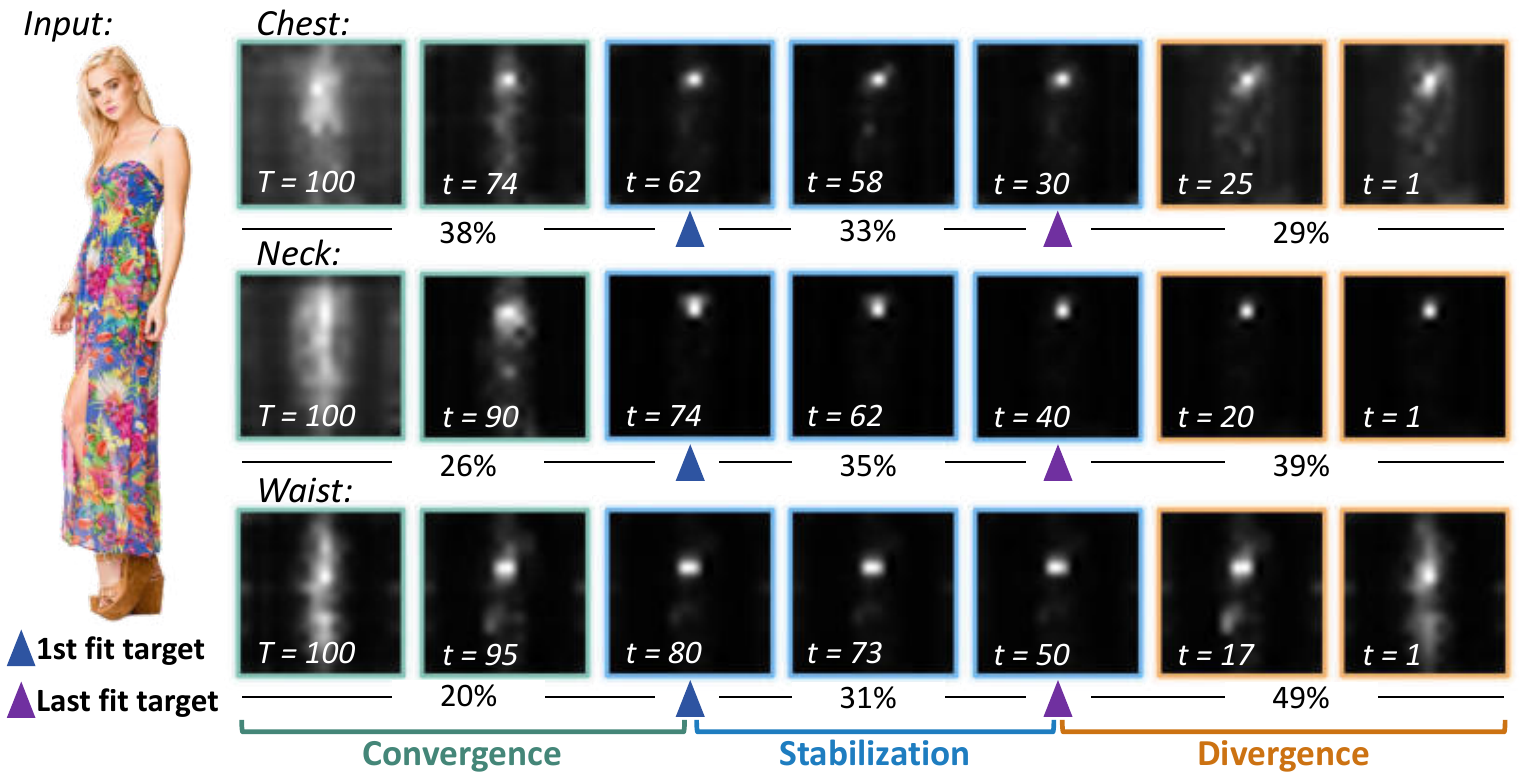}
   \caption{The spatial coverage of token attention maps evolves dynamically across diffusion steps, reflecting phase-specific localization patterns (Phase I/II/III). Details in Sec.\ref{C2-Region}.}
   \label{fig:eq3}
\end{figure}
\begin{itemize}
    \item[ ] \textbf{\textcolor[RGB]{68,134,120}{I Convergence:}}
 Attention converges from over-scattered to target-aligned regions (about $30\% \pm 15\%$ initial steps).
    \item[ ] \textbf{\textcolor[RGB]{30,125,192}{II Stabilization:}} Attention is steadily focused on target anatomical structures (about $30\% \pm 5\%$ middle steps).
    \item[ ] \textbf{\textcolor[RGB]{204,114,18}{III Divergence:}}
    Attention is either under- or over-shoots due to diffusion instability (about $30\% \pm 20\%$ final steps).
\end{itemize}
Core technical barriers due to dynamic attention are:
1) Dynamic Phase Conflicts: Adjacent regions (\emph{e.g., should vs. waist}) exhibit conflicting phases (I \emph{vs.} II) at the same step $t$.
2) Phase-Limited Noise: Early steps yield over-scattered attention (Phase I), while late steps suffer divergent predictions (Phase III).
3) Step-Averaging Artifacts: Naive cross-step aggregation amplifies boundary instability. 
To address these obstacles, we design Multi-Phase Token Filtering: 

\textbf{Thresholded Mask Averaging.}
\begin{figure}[t]
  \centering
   \includegraphics[width=0.97\linewidth]{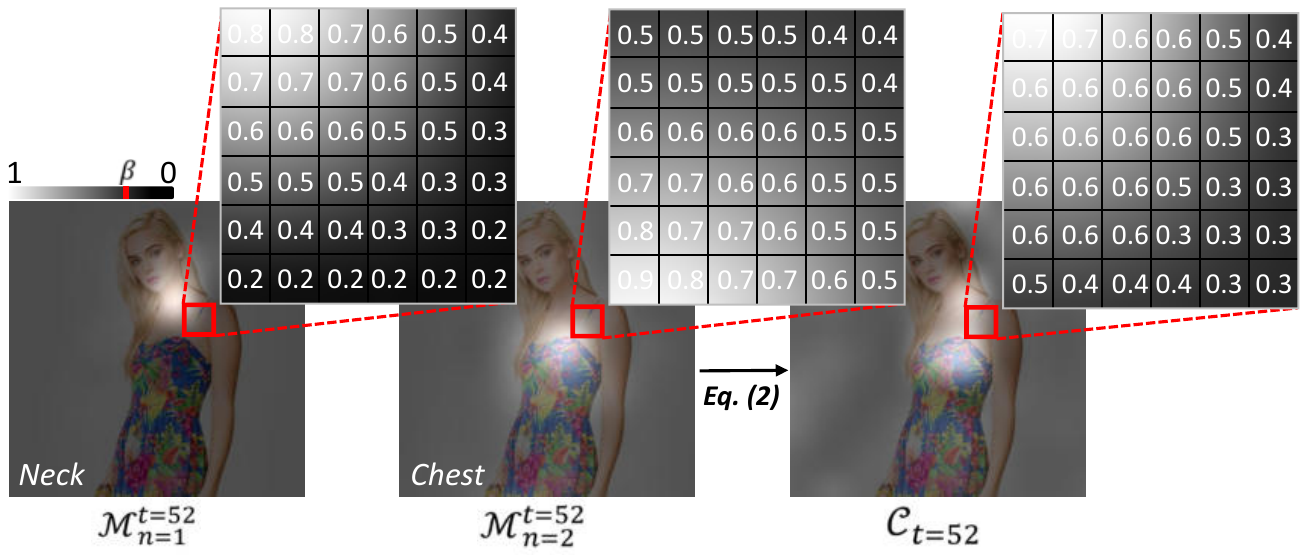}
   \caption{\textbf{Thresholded Mask Averaging} for Anatomical Token Fusion.
At diffusion step $t$, thresholded averaging merges neck and chest token maps (Eq.(\ref{eq:eq2})). Zoom in for details.}
   \label{fig:eq2}
\end{figure}
As shown in \cref{fig:eq2}, to address Dynamic Phase Conflicts and Phase-Limited Noise arising from \emph{single-step anatomical map fusion},
we selectively aggregates anatomically relevant regions while suppressing phase-specific noise.
For token map $\mathcal{M}_{t}   \in \mathbb{R} ^{N\times 16\times 16}$ at step $t$, we apply:
\begin{equation}
  \mathcal{C}_{t} = \frac{\sum_{n\in \left \{  1,...,N\right \} }^{}TH{\rm\textbf{-}}value\left ( \mathcal{M}_{n}^{t},\;\beta  \right )  }{TH{\rm\textbf{-}}count\left ( \mathcal{M}^{t},\;\beta  \right )},
  \label{eq:eq2}
\end{equation}
where 
$TH{\rm\textbf{-}}value(\cdot )$ filtering low confidence pixels based on $\beta$ threshold of $\beta = 0.3$.
$TH{\rm\textbf{-}}count(\cdot )$ is high-confidence count, $TH{\rm\textbf{-}}count(\cdot )\in [0,N]$.
This retains stable regions in Phase II while suppressing Phase I/III outliers, yielding coarse target maps $\mathcal{C}\in \mathbb{R} ^{100\times 16\times 16}$.

\textbf{Sliding Window Consensus.}
To address Step Averaging Artifacts and Phase-Limiting Noise arising from \emph{cross-step coarse target map fusion}, we propose phase-adaptive weighting.
Each coarse target map $\mathcal{C}_{t}$ is assigned a time-dependent weight $w_{t} = 2t/(T+1) $, where $T = 100$, $\sum_{t=1}^{T}w_{t} = 1$,  to progressively reduce unstable contributions from Phase III while preserving stabilized regions from Phase II. 
The reshaped tensor $\mathcal{C}\in \mathbb{R} ^{10\times 10\times 16\times 16}$  undergoes $3\times 3$ sliding window operations (no padding) to guide Phase II optimization Phase I/III.
\begin{equation}
  \hat{\mathcal{C}}_{i,j}  = \frac{\sum_{m=i}^{i+2}\sum_{n=j}^{j+2} \mathcal{C}_{m,n} \ast w_{m,n}}{\sum_{m=i}^{i+2}\sum_{n=j}^{j+2}w_{m,n}}  ,
  \label{eq:important}
\end{equation}
where $w_{t}$ reshape to $w_{m,n}$. 
Column-wise averaging of the refined $\hat{\mathcal{C}}\in \mathbb{R} ^{8\times8\times 16\times 16}$ produces fine target maps $\bar{\mathcal{C}}\in \mathbb{R} ^{8\times16\times 16}$.  
By harmonizing multi-phase evidence through spatially and temporally constrained fusion, the proposed method ensures robust anatomical localization even under extreme poses, fulfilling the core requirements derived from our earlier technical obstacles.

\subsection{Optimization: Edge-Aware Selection}\label{C3-Contour}
To enhance boundary sharpness while preserving anatomical topology, we propose merging cross- and self-attention maps—inspired by diffusion-based segmentation works (DiffuMask~\cite{Wu2023DiffuMaskSI} and DiffSeg~\cite{Shuai2024DiffSegAS}). Unlike prior methods~\cite{Wu2023DiffuMaskSI, Shuai2024DiffSegAS} that rely on self-attention for pixel-level classification, our analysis of DeepFashion~\cite{Ge2019DeepFashion2AV} samples reveals:
\begin{itemize}
    \item Cross-attention localizes fine-grained anatomy but suffers from blurred boundaries.
    \item Self-attention captures sharp edges but lacks anatomical specificity.
\end{itemize}
This complementary relationship motivates our Cross-Self Attention Merge, which synergizes the strengths of both mechanisms to achieve precise localization with crisp boundaries.

\textbf{Cross-Self Attention Merge.}
Overall, we employ pixel-level fusion for cross-attention and self-attention maps. This process averages corresponding pixels between target regions in cross-attention maps (which effectively localize areas of interest) and relevant regions in self-attention maps, using the former to mask irrelevant areas in the latter. The fused regions are subsequently filtered via thresholding. 
Specifically, we upsample the fine target map $\bar{\mathcal{C}}_{k}$ (cross-attention) to  $32\times 32$ and fuse $\bar{\mathcal{C}}\in \mathbb{R} ^{8\times32\times 32}$ with the final-step self-attention map $\mathcal{S}\in \mathbb{R} ^{8\times32\times 32}$ :
\begin{equation}
  \mathcal{R}_{k}   = TH{\rm\textbf{-}}value(\bar{\mathcal{C}}_{k} \oplus \mathcal{S}_{k},\;\alpha), \;  k\in\left \{ 0,...,7 \right \} ,
  \label{eq:important}
\end{equation}
$\oplus$ denotes pixel-wise averaging of cross-attention ($\bar{\mathcal{C}}_{k}$) and self-attention ($\mathcal{S}_{k}$) maps, followed by $TH{\rm\textbf{-}}value(\cdot )$ retains pixels ($>\alpha = 0.4$) to obtain fused regions $\mathcal{R}_{k}$. 

\textbf{Edge-Aware Selector.}
The initial mask $\mathcal{R}_{k}$ is obtained by rendering target regions ($>0$) of the attention map $\mathcal{R}_{k}$ as white and non-target regions ($=0$) as black.
To align boundaries of $\mathcal{R}_{k}$ with image content, we extract Canny edges $E$ from the input image $I$ and enforce geometric consistency:
\begin{equation}
\hat{E}(i,j) = \left\{\begin{matrix}
  1,& if \; E(i,j)=1,  \sqrt{(i-m)^{2} + (j-n)^{2}} \le \mu ,\\
  0,& otherwise .&
\end{matrix}\right.
  \label{eq:important}
\end{equation}
By defining a circular neighborhood (with center $\mathcal{R}_{k}(m,n)$, radius $\mu$) around each edge pixel $E(i,j)$, edges that align with anatomical contours are selectively preserved.
Canny-based geometric constraints ensure edited regions respect both anatomical structure and pose geometry, maintaining boundary-pose consistency.
The current $\hat{E}$ constitutes only an edge-aligned image. To convert $\hat{E}$ into a valid mask $\hat{M}$, we perform a series of post-processing operations (see Appendix \ref{Post-Processing}).
The generated human mask $\hat{M}$ seamlessly integrates with mask-guided editors (\emph{e.g.,} PowerPaint~\cite{Zhuang2023ATI}) without fine-tuning.


\section{Experiment}
\begin{figure*}[t]
  \centering
   \includegraphics[width=0.98\linewidth]{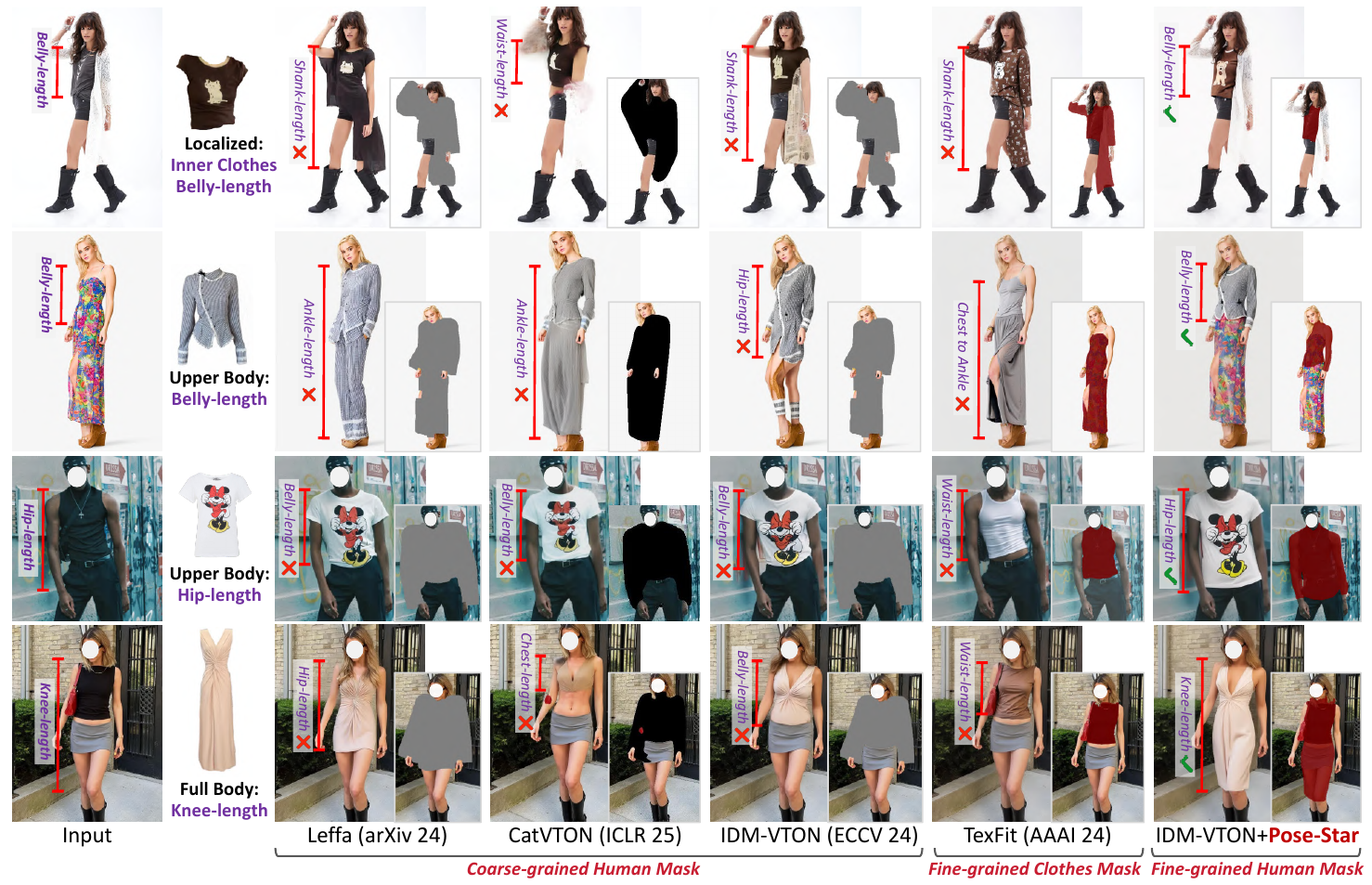}
   \caption{\textbf{Qualitative: Comparison with Fashion-specific Image Editors.} Includes state-of-the-art virtual try-on methods and a text-driven fashion image editing method called TexFit~\cite{Wang2024TexFitTF}.  Zoom in for details. Please see details in Sec.\ref{Effectiveness}.}
   \label{fig:C-1}
\end{figure*}

In this section, we comprehensively evaluate Pose-Star framework by addressing three critical research questions:
\begin{itemize}
    \item \textbf{Q1: Effectiveness.}
Does fine-grained human mask-guided editing outperform state-of-the-art fashion-specific and general-purpose image editors in both qualitative and quantitative comparisons?
    \item \textbf{Q2: Irreplaceability.}
Are the core components—Localization module, Aggregation module, Optimization module—irreplaceable for anatomically consistent detection of location, region, contour?
    \item \textbf{Q3: Robustness.}
How does the method perform under varying hyperparameters?
\end{itemize}
We answer \textbf{Q1} and \textbf{Q2} through systematic comparisons and ablation studies (Sec.\ref{Effectiveness} and Sec.\ref{Irreplaceability}), while sensitivity analysis for \textbf{Q3} is detailed in Appendix \ref{Robustness}.
The setting and evaluation datasets of comparison methods in Appendix \ref{Implementation}.

\subsection{Effectiveness}\label{Effectiveness}
This section addresses \textbf{Q1} by analyzing the performance of Pose-Star on three dimensions: User-defined region flexibility, pose robustness, and in-the-wild adaptability. 
Extended in-the-wild evaluations in the Appendix \ref{Wild Images}.

\textbf{Qualitative Comparison.}
Comparison with Fashion Editors is shown in \cref{fig:C-1}. 
Three key observations emerge: 
1) Flexible Anatomical Adaptation:
Unlike rigid clothing masks in TexFit~\cite{Wang2024TexFitTF}, which overfit clothes edges (\emph{e.g.,} preventing sleeve length changes), Pose-Star dynamically adjusts to user-defined anatomical regions (\emph{e.g., "short → long sleeves"}) through diffusion-based token maps (Sec.\ref{C1-Location}), enabling seamless style transformations without boundary artifacts.
2) Dynamic Region Localization:
Pose-Star overcomes the fixed anatomical priors of IDM-VTON~\cite{Choi2024ImprovingDM} and Leffa~\cite{Zhou2024LearningFF} (limited to predefined zones like \emph{"upper torso"}) by leveraging localization module (Sec.\ref{C1-Location}). This allows precise editing of layered clothes (\emph{e.g.,} modifying only the inner layer of a jacket in \cref{fig:C-1} Line1) through prompt-driven mask generation (\emph{e.g., "inner shirt"}).
3) Adaptive Length Control:
Existing methods fail to truncate full-length dresses to "belly-length" (\cref{fig:C-1} Line2) or extend "waist-length" tops to "hip-length" (\cref{fig:C-1} Line3) due to fixed mask boundaries. Aggregation module of Pose-Star (Sec.\ref{C2-Region}) fuses phase-specific attentions (convergence, stabilization, divergence), enabling anatomically grounded length adjustments while preserving pose coherence.
These advantages collectively resolve fundamental limitations of prior work, demonstrate the advantages of Pose-Star in user-defined flexibility, dynamic localization, and anatomical adaptability.

\begin{figure*}[h]
  \centering
   \includegraphics[width=0.97\linewidth]{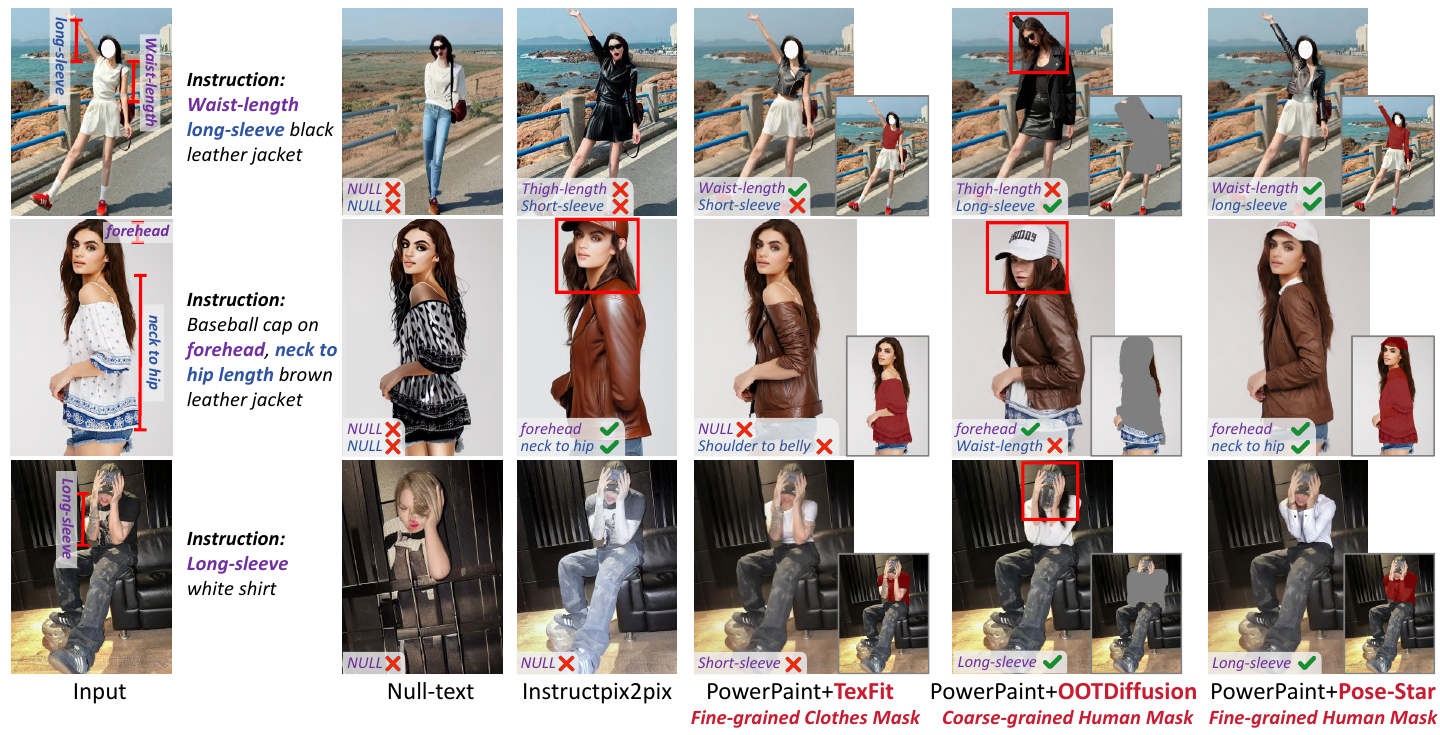}
   \caption{\textbf{Qualitative: Comparison with General-purpose Image Editors.} Includes two text-guided image editing methods and two image inpainting methods with different mask sources. Zoom in for details. Please see details in Sec.\ref{Effectiveness}.}
   \label{fig:C-2}
\end{figure*}
Comparison with General Editors is shown in \cref{fig:C-2}. 
Fine-grained clothes masks (PowerPaint + TexFit) will not work when altering styles (\emph{e.g., "T-shirt → long sleeve"}).
Coarse-grained human masks introduce identity distortion in complex poses (\emph{e.g., twisted torsos}).
Pose-Star combines optimization module (Sec.\ref{C3-Contour}) with pose-calibrated tokens to preserve facial identity and background details.

\begin{figure*}
  \centering
  \begin{subfigure}{0.7\linewidth}
\tabcolsep=0.2cm 
\resizebox{\linewidth}{!}{ 
\begin{tabular}{ccccc}
\toprule[0.4mm]
\textbf{Method}                                        & \textbf{Mask Type}                 & \textbf{User-Defined}$\uparrow$                                                            & \textbf{Pose Robustness}$\uparrow$                                                            & \textbf{In-the-Wild}$\uparrow$                                                             \\ \hline

Null-text~\cite{Null-text}                               & NULL                      & 1.03\footnotesize{$\pm$0.97}                           & 0.82\footnotesize{$\pm$0.31}                           & 2.78\footnotesize{$\pm$0.90}                           \\
Instructpix2pix~\cite{Brooks2022InstructPix2PixLT}                         & NULL                      & 3.57\footnotesize{$\pm$0.42}                           & 2.01\footnotesize{$\pm$0.10}                           & 3.96\footnotesize{$\pm$0.37}                           \\
TexFit~\cite{Wang2024TexFitTF}                                  & FCM & 1.22\footnotesize{$\pm$0.93}                           & 3.82\footnotesize{$\pm$0.33}                           & 3.00\footnotesize{$\pm$0.02}                           \\
Leffa~\cite{Zhou2024LearningFF}                                   & CHM & 2.93\footnotesize{$\pm$0.75}                           & 4.22\footnotesize{$\pm$0.40}                           & 3.86\footnotesize{$\pm$0.83}                           \\
CatVTON~\cite{Chong2024CatVTONCI}                                 & CHM & 1.73\footnotesize{$\pm$0.94}                           & 4.43\footnotesize{$\pm$0.07}                           & 4.10\footnotesize{$\pm$0.20}                           \\
PPt~\cite{Zhuang2023ATI}+TexFit                              & FCM & 1.31\footnotesize{$\pm$0.69}                           & 3.11\footnotesize{$\pm$0.93}                           & 4.23\footnotesize{$\pm$0.73}                           \\
PPt~\cite{Zhuang2023ATI}+OOTD~\cite{Xu2024OOTDiffusionOF}                                & CHM & 2.11\footnotesize{$\pm$0.52}                           & 1.62\footnotesize{$\pm$0.70}                           & 4.02\footnotesize{$\pm$0.41}                           \\
IDM~\cite{Choi2024ImprovingDM}+SCHP~\cite{Li2019SelfCorrectionFH}                                & CHM & 1.30\footnotesize{$\pm$0.43}                           & 3.78\footnotesize{$\pm$0.33}                           & 4.29\footnotesize{$\pm$0.03}                           \\ \hline
PPt~\cite{Zhuang2023ATI}\textbf{\textcolor[RGB]{112,48,160}{+Pose-Star}}\textbf{(Ours)} & \textbf{FHM}   & \textbf{4.85\footnotesize{$\pm$0.33}} & \textbf{4.74\footnotesize{$\pm$0.41}} & \textbf{4.63\footnotesize{$\pm$0.62}} \\
IDM~\cite{Choi2024ImprovingDM}\textbf{\textcolor[RGB]{112,48,160}{+Pose-Star}}\textbf{(Ours)} & \textbf{FHM}   & \textbf{4.89\footnotesize{$\pm$0.21}} & \textbf{4.82\footnotesize{$\pm$0.04}} & \textbf{4.55\footnotesize{$\pm$0.38}} \\ \toprule[0.4mm]
\end{tabular}
}
    \caption{The participants were asked to rate: (1) User-defined Region Flexibility, (2) Pose Robustness, and (3) In-the-Wild Generalizability. CHM denotes Coarse-grained Human Mask, FCM denotes Fine-grained Clothes Mask, and FHM denotes Fine-grained Human Mask. The perfect score is 5. Details in Appendix \ref{User Study}.}
    \label{fig:table}
  \end{subfigure}
  \hfill
  \begin{subfigure}{0.23\linewidth}
    \includegraphics[width=1\linewidth]{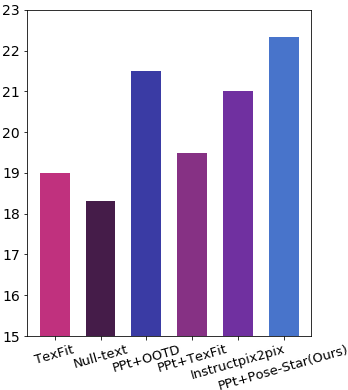}
    \caption{CLIP-Scores results of our method compared with text-guided image editing methods.}
    \label{fig:CLIP}
  \end{subfigure}
  \hspace{4mm}
  \caption{\textbf{Quantitative: }User feedback survey \cref{fig:table} and CLIP-Scores~\cite{Hessel2021CLIPScoreAR} comparison \cref{fig:CLIP}. PPt is PowerPaint~\cite{Zhuang2023ATI}, OOTD is OOTDiffusion~\cite{Xu2024OOTDiffusionOF}, IDM is IDM-VTON~\cite{Choi2024ImprovingDM}. }
  \label{fig:table-CLIP}
\end{figure*}

\textbf{Quantitative Comparison}
As shown in \cref{fig:table}, the integration of Pose-Star achieves significant improvements in pose robustness ($+1.81$ points), in-the-wild generalizability ($+0.81$ points), and user-defined flexibility ($+3.11$ points), uniquely advancing all three dimensions simultaneously—unattainable by prior methods.
Further, CLIP-Scores~\cite{Hessel2021CLIPScoreAR} were used to evaluate edited effects (\cref{fig:CLIP}).

\subsection{Irreplaceability}\label{Irreplaceability}
To address \textbf{Q2}, we conduct ablation studies by replacing each core component of Pose-Star with state-of-the-art alternatives while retaining other modules.
\begin{figure*}[h]
  \centering
   \includegraphics[width=0.98\linewidth]{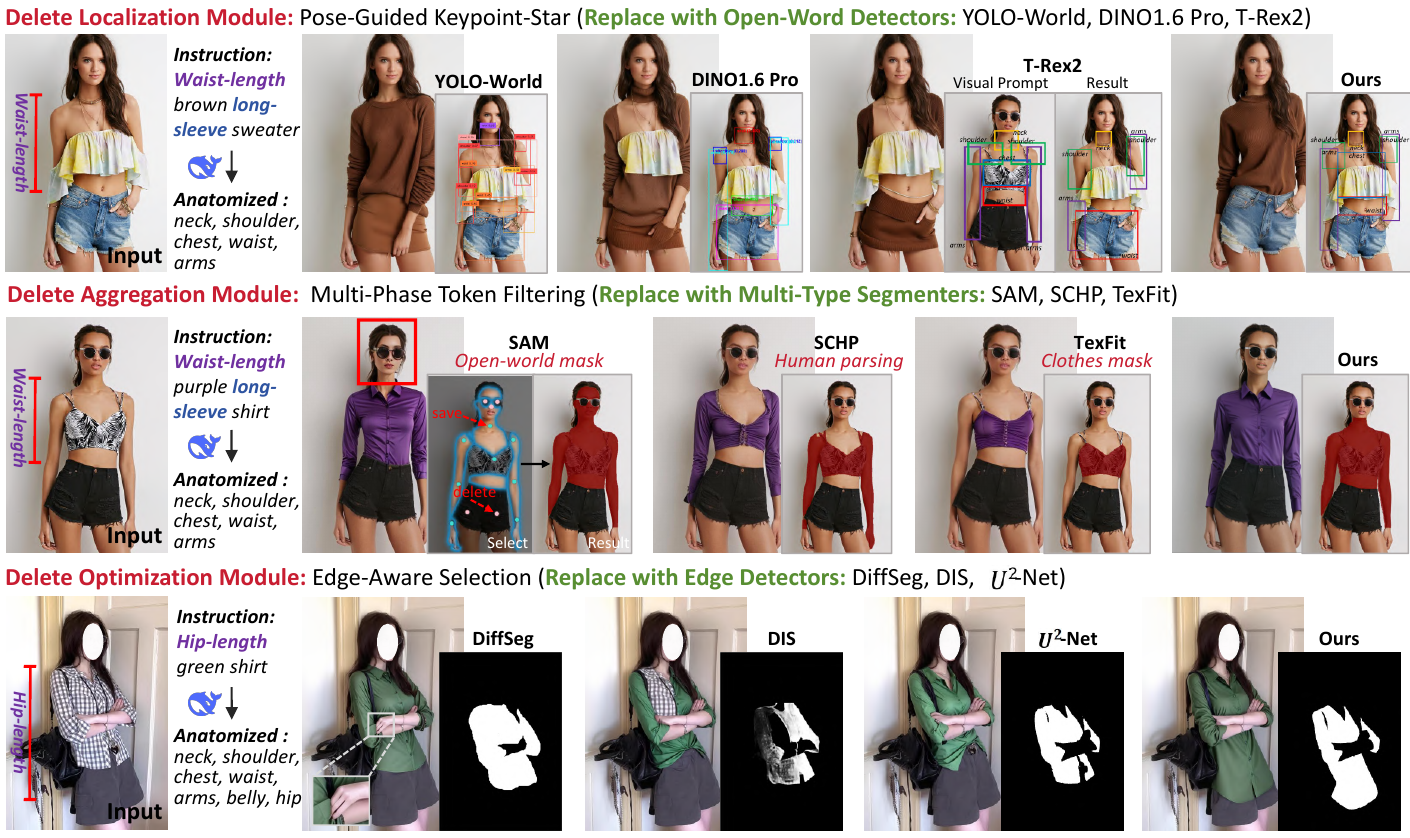}
   \caption{\textbf{Irreplaceability.} For each component, replace with the state-of-the-art tool, other components remain unchanged (See Sec.\ref{Irreplaceability}). 
   }
   \label{fig:C-3}
\end{figure*}

\textbf{Localization Module \emph{vs.} Open-Word Detectors:} 
Existing open-word detection methods struggle with anatomically precise editing due to three fundamental shortcomings: inability to localize uncommon anatomical regions (\emph{e.g.,} T-Rex2~\cite{Jiang2024TRex2TG} fails to detect chest, causing discontinuous edits), visual deceptive range in complex poses causes misclassification (\emph{e.g.,} DINO1.6 Pro~\cite{Ren2024GroundingD1} mistakes shorts for waist in complex poses), and off-target modifications (\emph{e.g.,} YOLO-World~\cite{Cheng2024YOLOWorldRO} erroneously edits jeans). In contrast, our localization module  (Sec.\ref{C1-Location}) addresses these issues through: 1) Diffusion-based token recognition leverages open-word capability of Stable Diffusion to detect rare anatomical regions; 2) Pose-driven token calibration aligns tokens with skeletal keypoints (OpenPose~\cite{Cao2018OpenPoseRM}) for robust localization in articulated poses; 3) Star-region constraints suppresses irrelevant regions via radial constraints. This synergistic approach achieves unprecedented editing precision, outperforming detection-centric methods in both anatomical consistency and pose robustness (\cref{fig:C-3} Line1).

\textbf{Aggregation Module \emph{vs.} Multi-Type Segmenters:}
SAM~\cite{Kirillov2023SegmentA} open-world segmentation via user clicks (triggered by localization centroids from Sec.\ref{C1-Location}). SAM expansion based on pixel-similarity causes uncontrolled over-segmentation (\emph{e.g., modifying non-target facial regions}), leading to identity loss.
SCHP~\cite{Li2019SelfCorrectionFH} fixed 20-class human parsing achieves high precision but lacks flexibility—it fails to segment localized regions like waist or neck.
TexFit~\cite{Wang2024TexFitTF} clothes-centric segmentation severely under-segments anatomical regions (\emph{e.g.,} $\ll50\%$ coverage for \emph{waist-length}), preventing structure-aware editing.
In contrast, our aggregation module (Sec.\ref{C2-Region}) preserves valid anatomical regions through phase-aware aggregation (convergence, stabilization, divergence), filters noise via thresholded averaging, and fits user-specified targets. This results in 62\% less boundary jitter than SAM and 92.5\% valid region retention (\emph{vs.} SCHP’s 48.2\%), demonstrating unparalleled precision for anatomical editing (\cref{fig:C-3} Line2).

\textbf{Optimization Module \emph{vs.} Edge Detectors:}
While DiffSeg~\cite{Shuai2024DiffSegAS} leverages diffusion priors for generalization, its coarse pixel-level classification generates noisy, anatomy-agnostic edges (\emph{e.g., distorted hands} in \cref{fig:C-3} Line3), failing to preserve fine anatomical contours. Similarly, DIS~\cite{Qin2022HighlyAD} and $U^{2}$-Net~\cite{Qin2020U2NetGD}, designed for foreground-background segmentation, erroneously classify subtle anatomical boundaries like hip-length as background, retaining only clothing edges that misalign with user intents. 
In contrast, our optimization module by Cross-Self Attention Merge and Edge-Aware Selector selectively refines image-aligned edges (\emph{e.g., clothes borders}) while preserving specific anatomy-specific regions (\emph{e.g., hip-lengths}), which skillfully harmonizes visual boundary and anatomical range fidelity.
This dual-mechanism approach reduces hand/foot distortions by 35\% compared to DIS/$U^{2}$-Net, validating its superiority in pose-aware edge optimization (\cref{fig:C-3} Line3).

\section{Conclusion}
Pose-Star addresses anatomy-aware mask generation by dynamically synthesizing fine-grained masks from semantic instructions in fashion editing. 
Integrated with existing editors, it significantly enhances key metrics like user-defined flexibility and pose robustness.
While diffusion inversion incurs computational overhead (3.6s/image on NVIDIA A100), subsequent anatomical edits require only 0.8s/task through optimized token aggregation. 
Future work could explore DPM-Solver++~\cite{Lu2022DPMSolverFS} for acceleration. Though imperfect, Pose-Star solves the overlooked problem of instruction-specific editing in articulated poses, spotlighting unresolved open-domain challenges. 
Discussion of limitations in Appendix \ref{Discussion}.

\section{Acknowledgments}
This work is supported by the National Natural Science Foundation of China under Grants (62176188),  the Innovative Research Group Project of Hubei Province under Grants (2024AFA017), the Major Project of Science and Technology Innovation of Hubei Province (2024BCA003) and the Key Project of Hubei Provincial Health Commission (WJ2023Z003).

{
    \small
    \bibliographystyle{ieeenat_fullname}
    \bibliography{main}
}

\appendix
\onecolumn

\section{Related Work}
\label{Related Work}
\paragraph{Existing Image Editing Methods}
Recent advancements in fashion image editing~\cite{Zeng2022FlowFaceSF, Chen2024ZeroshotIE, article, Li2025A} have enabled transformative applications such as virtual try-on~\cite{Choi2024ImprovingDM, Xu2024OOTDiffusionOF, Chong2024CatVTONCI, Chong2025CatV2TONTD}, dynamic garment replacement~\cite{jiang2025vidsketch, yang2025videograin}. Diffusion-based approaches, exemplified by VITON-HD~\cite{Choi2021VITONHDHV} (trained on 11,647 front-facing static poses), demonstrate high-fidelity garment synthesis but remain limited to controlled environments with simplified poses, hindering deployment in real-world scenarios. Predominant frameworks~\cite{Baldrati2023MultimodalGD, Gou2023TamingTP, Lee2022HighResolutionVT,
Lin2023FashionTexCV} adopt a two-stage paradigm: anatomical mask generation followed by diffusion-driven~\cite{Rombach2021HighResolutionIS} editing. However, mask generation strategies—whether coarse human parsers like SCHP~\cite{Li2019SelfCorrectionFH} or garment-specific segmenters such as TexFit~\cite{Wang2024TexFitTF} (encoders and decoders trained with the fashion-specific dataset DeepFashion)—struggle to reconcile pixel-level anatomical accuracy with open-vocabulary editing flexibility.

Coarse-grained human masks, exemplified by IDM-VTON~\cite{Choi2024ImprovingDM} (DensePose~\cite{Cao2018OpenPoseRM}-driven torso segmentation) and OOTDiffusion~\cite{Xu2024OOTDiffusionOF} (SCHP-based~\cite{Li2019SelfCorrectionFH} full-body parsing), restrict edits to predefined zones (\emph{e.g., upper torso}) through rigid anatomical priors, preventing dynamic length customization (\emph{e.g., waist-to-hip transformations}). Recent methods like CatVTON~\cite{Chong2024CatVTONCI} and Cat2VTON~\cite{Chong2025CatV2TONTD} expand editable regions via global attention mechanisms but retain lower-body constraints from ankle-length dress training data in VITON-HD~\cite{Choi2021VITONHDHV}. Conversely, Fine-grained clothes masks (DPDEdit~\cite{wang2024dpdedit} with Grounded-SAM~\cite{Kirillov2023SegmentA} architecture, FICE~\cite{Pernu2023FICETF} via CLIP-guided grounding) achieve pixel-aligned boundaries but rigidly adhere to dataset-specific categories, making open-vocabulary style changes (\emph{e.g., "dress→pants"}) infeasible. Even state-of-the-art approaches like Leffa~\cite{Zhou2024LearningFF}, which introduces flow-guided attention for lower-body edits, cannot modify garment lengths due to fixed mask topologies. General-purpose editors~\cite{Li2024BrushEditAI, Ju2024BrushNetAP, Feng2024AnII} (\emph{e.g.,} Instructpix2pix~\cite{Brooks2022InstructPix2PixLT}, Null-text~\cite{Null-text}) suffer from color bleeding and detail loss (\emph{e.g., distorted hands}) due to unconstrained attention maps.
Emerging solutions like PromptDresser~\cite{Kim2024PromptDresserIT} explore adjustable masks for wrinkle editing but lack anatomical length control, underscoring a persistent gap: no existing method enables dynamic, anatomy-aware masking for open-world fashion editing.

\paragraph{Existing Detection/Segmentation Methods}
The quest for anatomically precise open-world editing reveals critical flaws in existing methods. Open-vocabulary detectors like T-Rex2~\cite{Jiang2024TRex2TG} (noun-centric) struggle to localize rare regions (\emph{e.g., chest}), while DINOv1.6 Pro~\cite{Ren2024GroundingD1} misclassifies ambiguous areas (\emph{e.g., shorts→waist in twisted poses}), and YOLO-World~\cite{Cheng2024YOLOWorldRO} erroneously edits non-target regions (\emph{e.g., jeans}). Segmentation tools like SAM~\cite{Kirillov2023SegmentA} suffer uncontrolled over-segmentation via pixel-similarity expansion (\emph{e.g., facial edits during torso adjustments}), distorting identity. Human parsers (20-class of SCHP~\cite{Li2019SelfCorrectionFH}, DeepFashions~\cite{Xiao2017FashionMNISTAN}-trained garment constraints of TexFit~\cite{Wang2024TexFitTF}) lack flexibility, omitting regions like waist or under-segmenting anatomy.

To achieve cross-modal alignment~\cite{Jiang_2023_CVPR,Synergy,Cao2023AnES,Cross-Modality}, diffusion-based~\cite{Rombach2021HighResolutionIS} approaches (DiffSeg~\cite{Shuai2024DiffSegAS}) generate noisy, anatomy-agnostic edges (\emph{e.g., distorted hands}) via coarse pixel classification, prioritizing texture over structure. Foreground-background segmenters (DIS~\cite{Qin2022HighlyAD}/$U^{2}$-Net~\cite{Qin2020U2NetGD}) misclassify subtle boundaries (hip-length→background), retaining only misaligned clothing edges. Early attempts to extract anatomy-aware masks (\emph{e.g.,} $U^{2}$-Net~\cite{Qin2020U2NetGD}/DIS~\cite{Qin2022HighlyAD} for belly-length regions) collapsed under foreground-background dichotomies, while DiffSeg’s text-aligned masks via inversion introduced unstable attention noise. Subsequent open-vocabulary detectors (GLEE~\cite{Wu2023GeneralOF}/T-Rex2~\cite{Jiang2024TRex2TG}), trained on noun-centric tags, failed catastrophically on rare anatomical prompts ((\emph{e.g., waist/belly}) and articulated poses due to unconstrained region proposals. Human parsers (M2FP~\cite{Yang2023DeepLT}/AIParsing~\cite{Zhang2022AIParsingAI}) further highlighted field-wide rigidity, omitting critical regions (\emph{e.g., chest/belly}). Collectively, these efforts expose a persistent gap: no method achieves user-defined anatomical segmentation with pose robustness and in-the-wild generalization, demanding dynamic mask generation that fuses diffusion’s open-vocabulary capacity with biomechanical priors.

\section{DeepSeek Anatomy Parser}
\label{DeepSeek}
\begin{table*}[h]
\renewcommand{\arraystretch}{1.1}
\centering
\setlength{\tabcolsep}{2.5mm}{
\begin{tabular}{c|cc}
\toprule
\textbf{Category}      & \textbf{Included Tokens}                                   & \textbf{Coverage Principle}            \\ \hline
\textbf{Star Tokens}   & Neck, Shoulder, Elbow, Wrist, Hip, Knee, Ankle             & Skeletal joints for pose calibration   \\
\textbf{Fleshy Tokens} & Forehead, Chest, Waist, Belly, Arms, Hip, Hand, Thigh, etc & Soft-tissue regions for volume editing \\ \bottomrule
\end{tabular}

}
  \caption{
   Star tokens (skeletal joints) and fleshy tokens (volumetric anatomy) classification rules. }
  \label{tab:tokens}
\end{table*}

\begin{table*}[h]
\renewcommand{\arraystretch}{1.1}
\centering
\setlength{\tabcolsep}{5mm}{
\begin{tabular}{c|ccc}
\toprule
\textbf{Garment}      & \textbf{Start Point} & \textbf{End Point} & \textbf{Coverage Zone}     \\ \hline
\textbf{Blouse/Shirt} & Neck/User-defined                 & User-defined       & Upper body (arms included) \\
\textbf{Dress}        & Neck/User-defined    & User-defined       & Full torso + legs          \\
\textbf{Pants/Skirt}  & Waist                & User-defined       & Lower body                 \\ \bottomrule
\end{tabular}

}
  \caption{
  \textbf{Partial Instruction Mapping Protocol.} Length Anchor: Explicit endpoint (\emph{e.g., belly in "belly-length"}), Implicit Start: Auto-derived from garment type.}
  \label{tab:instruction}
\end{table*}

\begin{figure*}[h]
  \centering
   \includegraphics[width=1\linewidth]{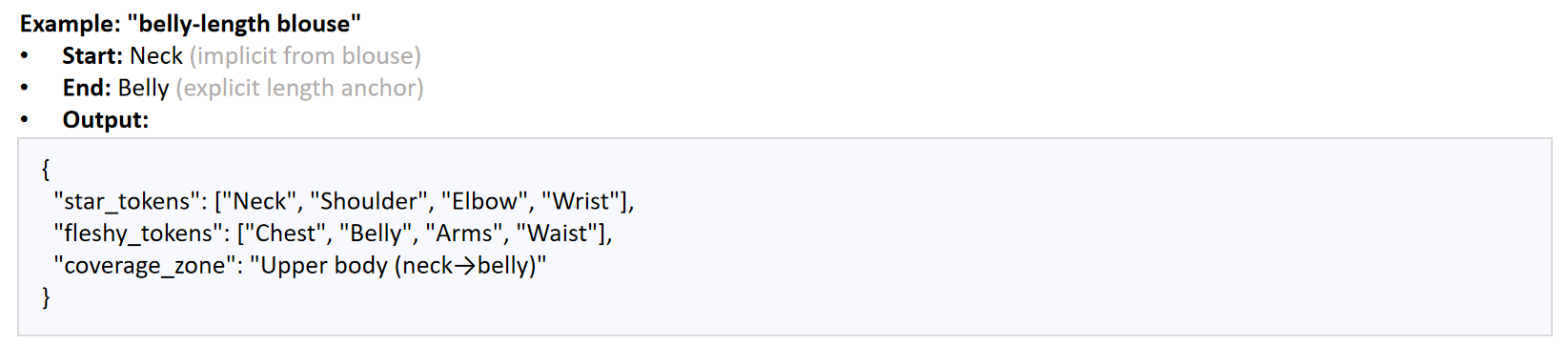}
   \caption{Instruction mapping protocol example and DeepSeek target output in .json form.}
   \label{fig:example}
\end{figure*}
The Pose-Star framework employs DeepSeek-V3-Base~\cite{DeepSeekAI2024DeepSeekV3TR}, which is based on the MoE architecture, as the anatomical instruction parser. This module interprets user-provided semantic commands (\emph{e.g., "belly-length blouse"}) into structured body representations that define garment coverage zones. As shown in \cref{tab:instruction},  the parser first identifies explicit length anchors (\emph{e.g., "belly"}) and implicit coverage ranges derived from garment semantics: blouses/shirts default to neck-to-hip coverage including arms; dresses extend from neck to user-specified endpoints; pants/skirts initiate from waist landmarks.
As shown in \cref{tab:tokens},  key to this process is the decomposition into two complementary anatomical token sets:
Star Tokens encode skeletal joints (Neck, Shoulder, Elbow, Wrist, Hip, Knee, Ankle) for pose calibration.
Fleshy Region Tokens represent volumetric anatomy (Chest, Waist, Belly, Arms, Thighs, Shanks, Torso) for style editing. 
A example of an anatomy-aware instruction parsing and output results is shown in \cref{fig:example}.

\section{Attention Mechanisms in U-Net Architecture}
\label{Attention}
\begin{figure*}[h]
  \centering
   \includegraphics[width=1\linewidth]{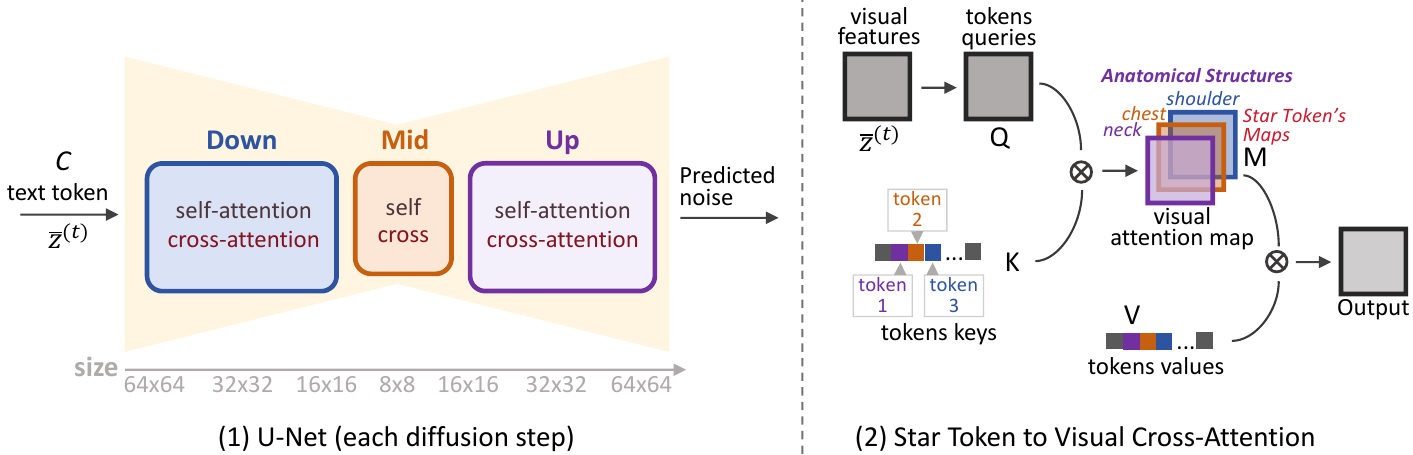}
   \caption{\textbf{Background and Preliminaries.} Diffusion-based token initialization.}
   \label{fig:Background}
\end{figure*}
To obtain the attention maps from the diffusion process of real images for inference in Pose-Star, we need to perform diffusion (i.e., inversion-reconstruction) on these real images. Specifically, we invert (add noise to) the real image into the initial latent space using DDIM~\cite{Song2020DenoisingDI} and then reconstruct (denoise) it back to the real image. This work primarily focuses on the reconstruction process, with the goal of acquiring attention maps that capture the text token-image mapping during this reconstruction. Additionally, due to the approximation inherent in diffusion models—where the U-Net cannot perfectly learn the theoretical reverse conditional distribution $q\left ( z_{t-1}\mid z_{t} , z_{0}  \right )$ and instead provides an approximate $p_{\theta } \left ( z_{t-1}\mid z_{t} \right ) $—the inversion-reconstruction process lacks perfect symmetry, leading to errors where the reconstruction struggles to fully restore the original image. To address this, we consider introducing null-text optimization~\cite{Null-text}. Its core idea is to align the reconstruction process with the inversion process by optimizing the MSE loss between the latent vector $\bar{z} _{t}$ from the reconstruction process and the corresponding latent vector $z_{t}$ from the inversion process at the same diffusion timestep:
\begin{equation}
  \min_{\emptyset_{t}  } \left \| z_{t-1} - z_{t-1}\left ( \bar{z}_{t}, \emptyset_{t}, C  \right ) \right \|_{2}^{2}, \quad
  \bar{z} _{t-1}=z_{t-1}\left ( \bar{z}_{t}, \emptyset_{t}, C  \right ). 
  \label{eq:important}
\end{equation}
Here, $\emptyset$ denotes the null text embedding, whose value is adjusted during optimization to minimize the loss. $C$ represents the text condition input, corresponding to target tokens (Star Tokens/Fleshy Tokens/Clothes Tokens), and $z_{t-1}\left ( \cdot  \right )$ signifies the latent vector update for a single diffusion step. Through this process, we obtain the latent diffusion trajectory of the real input image, from which we then extract the attention maps corresponding to the target tokens.

The U-Net structure used for noise prediction at each diffusion step is illustrated in \cref{fig:Background}(1), with our focus on its attention layer. This layer comprises a downsampling (down) block, a middle (mid) connection block, and an upsampling (up) block. Each block contains cross-attention and self-attention maps at different spatial resolutions: the down and up blocks include resolutions of 256, 1024, and 4096, while the mid block has a resolution of 64. Higher resolutions typically store higher-dimensional feature information. Crucially, as demonstrated in Prompt-to-Prompt~\cite{Hertz2022PrompttoPromptIE} and shown in Figure \cref{fig:Background}(2), the cross-attention layer primarily maps the text condition $C$ to a set of visual attention maps $M$, where each map in $M$ corresponds to a specific text token (\emph{e.g., 'neck'}). This finding is essential to our method. Specifically, the deep spatial features of the noisy image $\bar{z} _{t}$ are projected into the query matrix $Q$, while the text embeddings (including: Star Tokens/Fleshy Tokens/Clothes Tokens) are projected into the key matrix $K$ and value matrix $V$. Learned linear projections then yield the visual attention map corresponding to each token within the text condition $C$:
\begin{equation}
  M = Softmax\left ( \frac{QK^{T} }{\sqrt{d} }  \right ) .
  \label{eq:important}
\end{equation}
Through this method, the text condition $C$ is mapped to the output image $z_{0}$, ensuring it aligns with the given conditional prompt. We observe that the intermediate attention map set $M$ serves as a visual identifier for the spatial extent of each token, exhibiting a one-to-one correspondence with every text token. In essence, $M$ functions similarly to a detection result capturing the regions associated with all text tokens. By extracting the corresponding attention map from $M$, we initialize the Maps for Star Tokens, Fleshy Tokens, and Clothes Tokens.

\section{Post-Processing of Edge-Aware Selector.}
\label{Post-Processing}
To convert the refined edge map $\hat{E}$ into a valid segmentation mask, we perform a series of post-processing operations including Edge Discontinuity Optimization, Edge-to-Mask Conversion, and Mask Edge Smoothing. 
\paragraph{Edge Discontinuity Optimization}
We enforce spatial continuity in $\hat{E}$ to establish well-defined boundaries. Although the initial Canny edges $E$ suffer from fragmentation and missing segments due to illumination variations and interfering objects, we preserve relevant edges from $R_k$ while addressing discontinuities. Specifically, we bridge discontinuous endpoints through direct linear interpolation, as empirical evaluations indicate this approach introduces boundary deviations within 5\% of the total target area - an acceptable tolerance for practical applications. This optimization process ultimately yields a topologically continuous edge representation suitable for mask generation.
\paragraph{Edge-to-Mask Conversion}
To generate a binary mask $\hat{M}$ from a closed curve represented in $\hat{E}$ (where 1 indicates edge pixels and 0 indicates non-edge pixels), we propose an efficient boundary propagation algorithm that fills the interior region while preserving the curve topology. First, initialize $\hat{M}$ by copying $\hat{E}$ such that edge pixels retain value 1 and non-edge pixels are set to 0. Next, identify external regions by propagating from image boundaries: Enqueue all boundary-adjacent pixels $(i,j)$ where $\hat{M}_{i,j} = 0$ (non-edge), temporarily marking them as external (value 2). Perform breadth-first search using 4-connectivity (up/down/left/right neighbors), iteratively marking connected non-edge pixels as external (2). Upon queue exhaustion, finalize $\hat{M}$ by assigning 0 to external pixels (value 2), 1 to all remaining non-edge pixels (interior), and preserving original edge pixels (1). This approach robustly distinguishes interior/exterior regions in $O(HW)$ time while handling complex curve topologies through boundary-connected propagation, ensuring closed curves yield watertight masks.
\paragraph{Mask Edge Smoothing}
To address potential artifacts such as dark seams along mask boundaries that may arise from excessively precise segmentation during editing, we implement a Mask Edge Smoothing procedure to enhance coherence between edited and unedited regions. This process incorporates two complementary operations: First, we apply morphological dilation using a small circular kernel (radius=2 pixels) to slightly expand the mask boundary, ensuring sufficient coverage of transitional edge areas. Second, we employ Gaussian smoothing ($\sigma$=1.5) followed by curvature-constrained B-spline fitting to maintain geometric continuity while eliminating irregular jagged artifacts along the contour. These operations collectively preserve topological integrity while generating perceptually natural transitions, ultimately producing the optimized mask $\hat{M}$ that robustly supports seamless content generation in practical editing scenarios.


\section{Implementation}
\label{Implementation}
We systematically evaluate Pose-Star against state-of-the-art fashion-specific and general-purpose image editors:
1) Fashion-Specific Editors: IDM-VTON~\cite{Choi2024ImprovingDM} and CatVTON~\cite{Chong2024CatVTONCI} for virtual try-on based on coarse-grained human masks, Leffa~\cite{Zhou2024LearningFF} for full-body garment replacement using fixed anatomical priors, and TexFit~\cite{Wang2024TexFitTF} for text-driven editing with fine-grained clothes masks.
2) General-Purpose Editors: text-guided diffusion editors Null-text~\cite{Null-text} and InstructPix2Pix~\cite{Brooks2022InstructPix2PixLT} without masks, PowerPaint~\cite{Zhuang2023ATI} + TexFit~\cite{Wang2024TexFitTF}/OOTDiffusion~\cite{Xu2024OOTDiffusionOF} for inpainting with clothes/human masks.
To validate the effectiveness of Pose-Star's plug-in to the existing editor, our framework is configured for virtual try-on: IDM-VTON~\cite{Choi2024ImprovingDM}+Pose-Star, text-driven editing: PowerPaint~\cite{Zhuang2023ATI}+Pose-Star.
Pose-Star generates fine-grained human masks.
All experiments utilize the stable-diffusion-v1-4 model \footnote{https://huggingface.co/CompVis/stable-diffusion-v1-4} and are conducted on NVIDIA GeForce RTX 4090 GPUs with 24GB VRAM.
\begin{figure*}[h]
  \centering
   \includegraphics[width=1\linewidth]{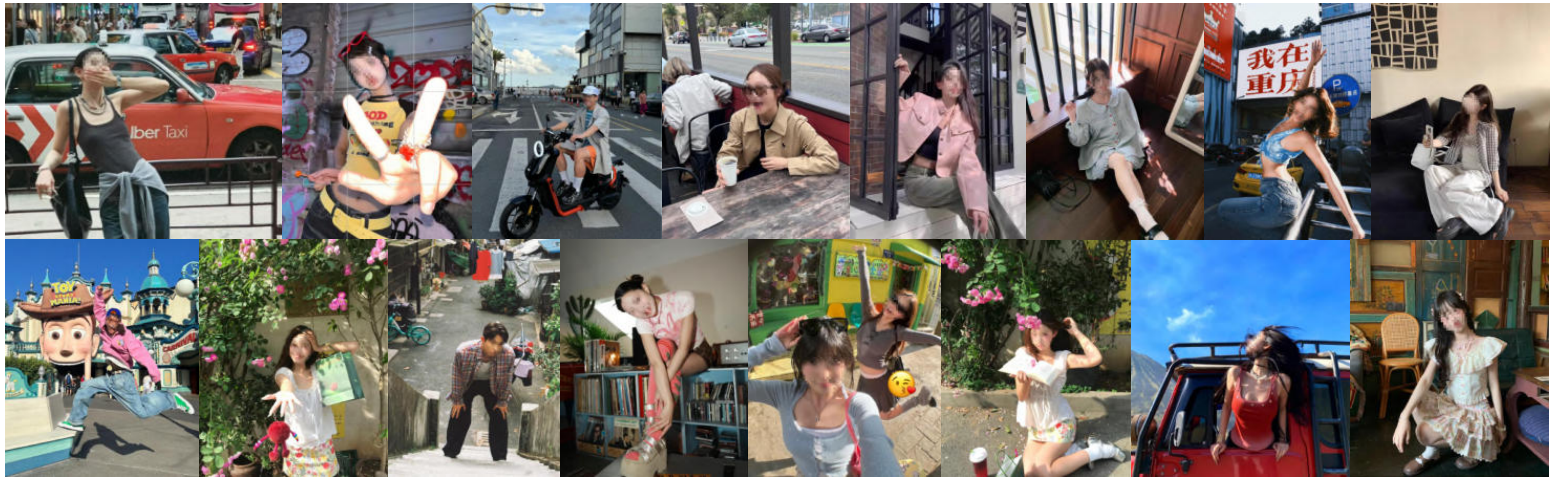}
   \caption{\textbf{Representative Samples from the Challenging Real-World Evaluation Dataset.}}
   \label{fig:Dataset}
\end{figure*}

As our approach is training-free, it eliminates the need for collecting large-scale, labor-intensive training datasets. To evaluate its practical utility in real-world, challenging scenarios, we curated 16,800 captured data samples from authentic online platforms and applications (Facebook, Xiaohongshu, YouTube frames). This dataset (as shown in \cref{fig:Dataset}) was meticulously selected to include challenging cases across multiple dimensions: hinged poses, dynamic snapshots, wide-angle overhead shots, layer occlusions, and diverse body types/ages. Additionally, our test suite incorporates 5,136 hinged pose samples filtered from DeepFashion-MultiModal. We will publicly release this challenging benchmark to better assess the limits of current model capabilities and enable targeted optimization.

\section{Robustness}
\label{Robustness}
\begin{figure*}[h]
  \centering
   \includegraphics[width=1\linewidth]{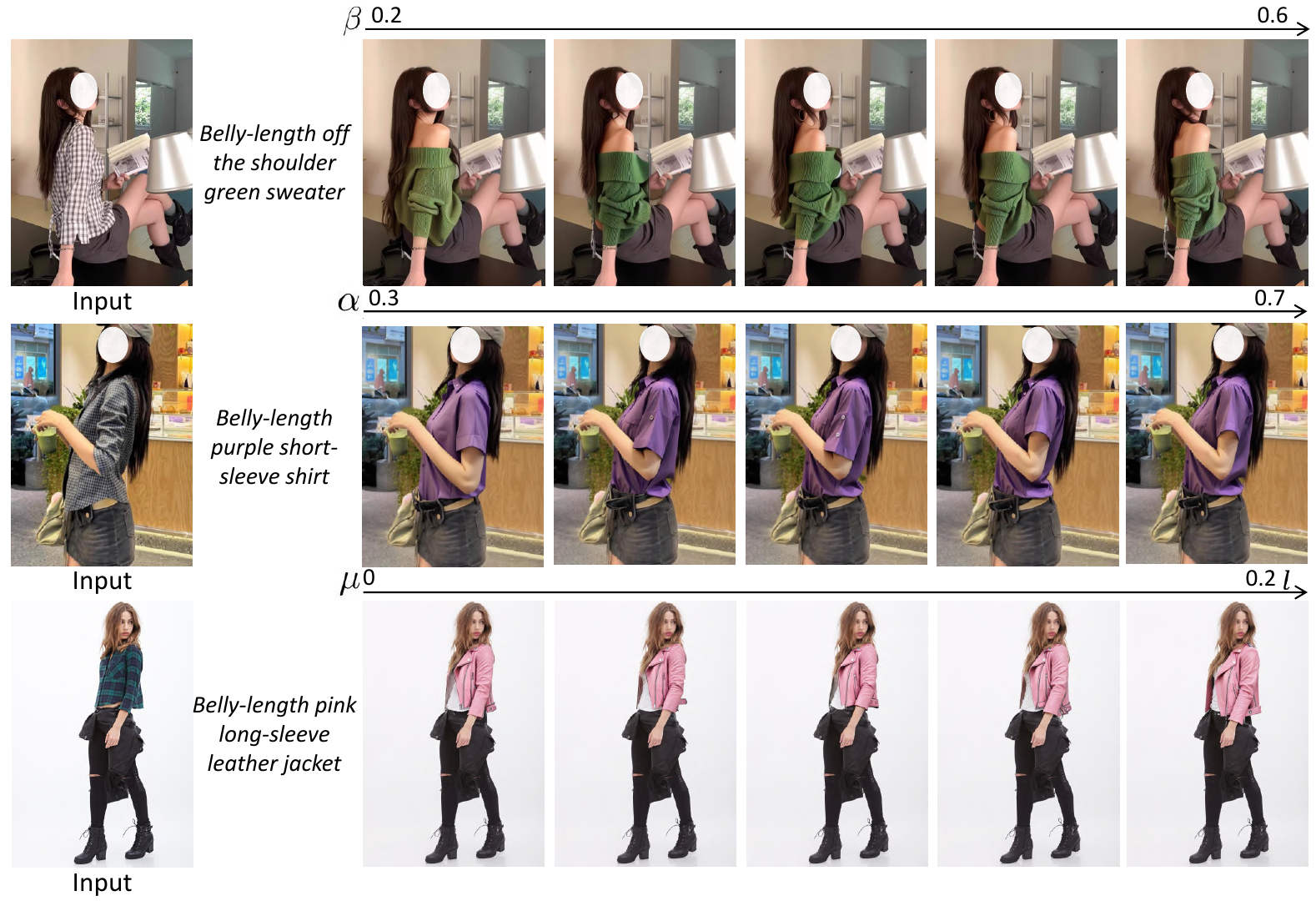}
   \caption{\textbf{Hyperparametric Evaluation.} }
   \label{fig:Sen}
\end{figure*}

This section addresses \textbf{Q3} by analyzing the performance of Pose-Star under varying hyperparameters. We focus on three critical parameters: threshold $\beta \in \left ( 0,1 \right )$  for thresholded mask averaging (Sec.\ref{C2-Region}), where higher $\beta$ enforces stricter attention filtering; threshold $\alpha  \in \left ( 0, 1 \right )$  for cross-self attention merge (Sec.\ref{C3-Contour}), with larger $\alpha$ tightening boundary alignment; and selection range $\mu   \in \left [ 0,1 \right ] $ for the edge-aware selector, where $l$ denotes the shortest distance from point  $\mathcal{R}_{k}(m,n)$ to boundary of $\mathcal{R}_{k}$, and smaller $\mu$ imposes stricter edge constraints. Our suggested setting range is: $\beta \in \left ( 0.2, 0.6 \right )$, $\alpha  \in \left ( 0.3, 0.7 \right )$, and $\mu \in \left ( 0, 0.2l \right ) $, as shown in \cref{fig:Sen}.

The analysis demonstrates that parameters within our recommended ranges effectively balance noise suppression and precision. Threshold $\beta$ provides first-stage noise filtering during region aggregation, while stricter thresholds $\alpha$ enhance second-stage refinement, retaining only anatomically coherent regions. For edge optimization, tighter $\mu$ values—enabled by the localization and fusion module’s inherent precision—yield sharper boundaries aligned with anatomical contours. Collectively, Pose-Star exhibits stable performance across these ranges, demonstrating robustness against parameter variations in open-world editing scenarios.

We further evaluate the impact of two key parameters: the radius $r$ for filtering attention noise around keypoints and the sliding window size for merging multi-stage token attention maps. For star-region constraints, $r$ is set to the minimum, maximum, and average. The sliding window size is evaluated across four configurations: 1 $\times$ 1, 2 $\times$ 2, 3 $\times$ 3, 4 $\times$ 4. Performance is measured using mean Intersection over Union (IoU), with quantitative results summarized in \cref{tab:Parametric}.
\begin{table*}[h]
\renewcommand{\arraystretch}{1.1}
\centering
\setlength{\tabcolsep}{3mm}{
\begin{tabular}{c|ccc|cccc}
\toprule
\textbf{Parameter Types} & \multicolumn{3}{c|}{\textbf{Radius $r$}}     & \multicolumn{4}{c}{\textbf{Window Size}}                  \\ \hline
\textbf{Settings}        & \textbf{min $r$} & \textbf{ave $r$} & \textbf{max $r$} & \textbf{1 $\times$ 1} & \textbf{2 $\times$ 2} & \textbf{3 $\times$ 3} & \textbf{4 $\times$ 4} \\ \hline
\textbf{Average IoU}~\cite{Rezatofighi2019GeneralizedIO}     & 0.69         & 0.73         & 0.67         & 0.51         & 0.67         & 0.73         & 0.69         \\ \bottomrule
\end{tabular}

}
  \caption{
  Parametric evaluation of Radius $r$ and Window Size. }
  \label{tab:Parametric}
\end{table*}

Based on the evaluation results, we observe that extremely small or large radius $r$ values lead to either loss of valid regions or excessive noise inclusion. Consequently, setting $r$ to the average distance achieves an optimal trade-off in stability. For sliding window size, results align with findings in Sec.\ref{C2-Region}: multi-stage attention map fusion (3 $\times$ 3) outperforms single-stage configurations (1 $\times$ 1, 2 $\times$ 2) by better balancing effective region coverage, while oversized windows (4 $\times$ 4) induce performance degradation due to over-coarsened fusion.

\begin{figure*}[h]
  \centering
   \includegraphics[width=1\linewidth]{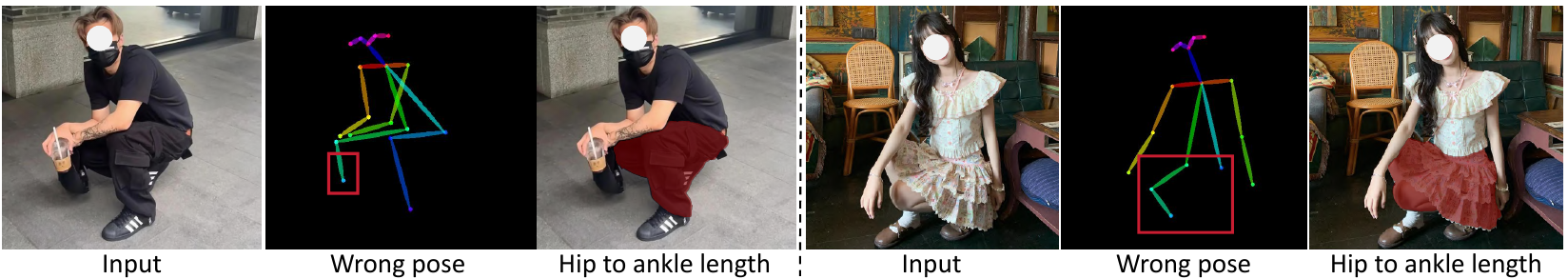}
   \caption{\textbf{Pose-Star Robustness Evaluation on OpenPose Partial Keypoints Errors.} }
   \label{fig:Keypoints}
\end{figure*}

Additionally, to evaluate our method's dependency on the pre-trained OpenPose model, we assess samples containing OpenPose keypoint detection errors. As illustrated in \cref{fig:Keypoints}, OpenPose frequently exhibits local keypoint misalignment or loss when the human body is heavily occluded by clothing (\emph{e.g., skirts}) or in the presence of similarly colored adjacent regions (\emph{e.g., black hats and trousers}). Nevertheless, Pose-Star reliably generates anatomically consistent masks despite keypoint inaccuracies. This robustness stems from OpenPose keypoints solely filtering attention pixels rather than directly generating attentions (Sec.\ref{C1-Location}), thus avoiding fundamental localization interference. Furthermore, clothing tokens mitigate the impact of erroneous keypoints, while calibration through aggregation and refinement modules ensures stable performance. Consequently, Pose-Star achieves resilience by synergizing existing components rather than relying on any single module. The framework’s modular design also supports alternative pose estimators, which we will explore in future work.

\section{User Study Protocol}
\label{User Study}
\begin{figure*}[h]
  \centering
   \includegraphics[width=0.9\linewidth]{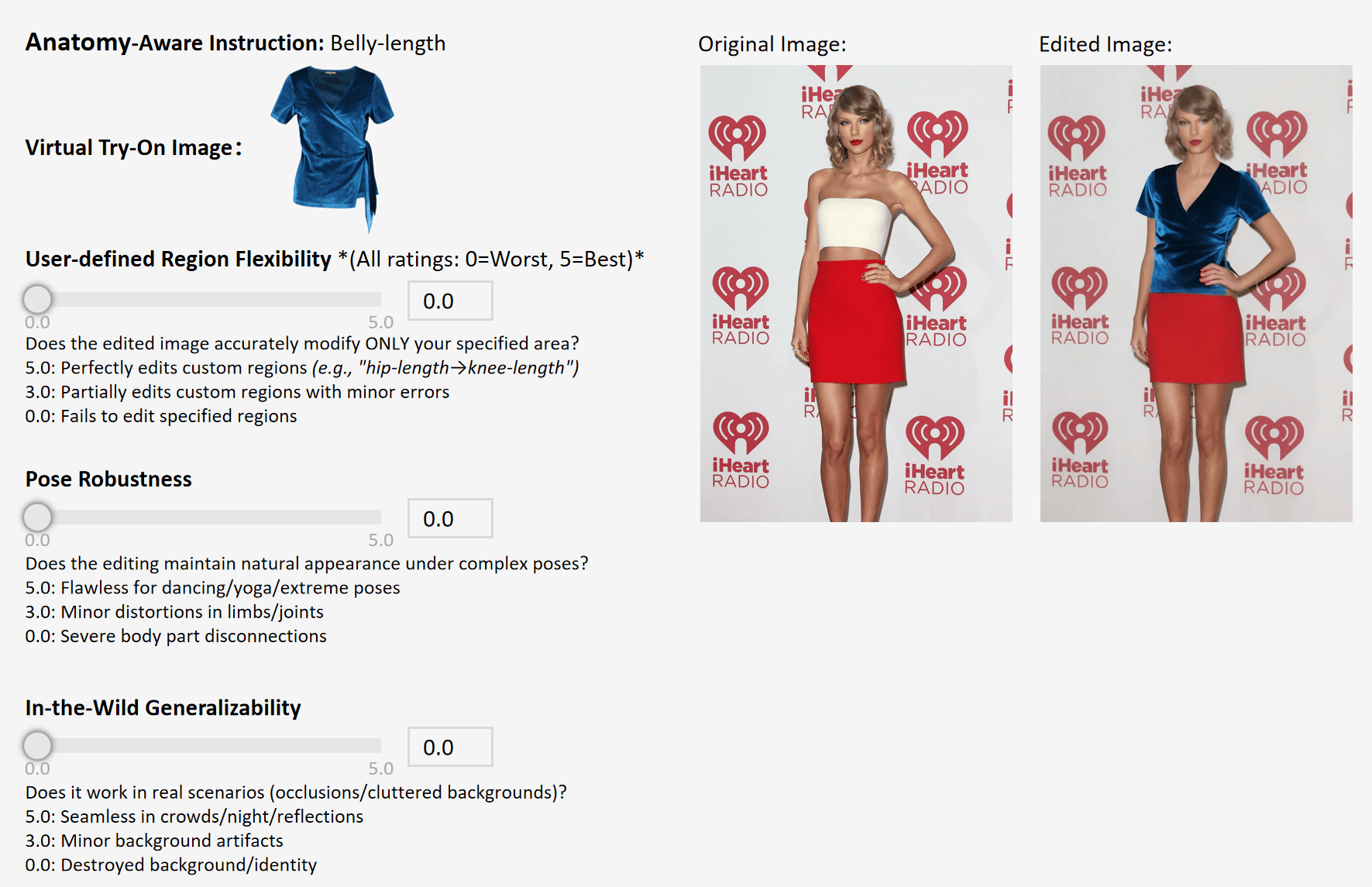}
   \caption{\textbf{ Participant Evaluation Interface.} Participants rate edited images across three dimensions: User-defined Region Flexibility, Pose Robustness, In-the-Wild Generalizability. Scores range 0–5 (5=Best).}
   \label{fig:survey}
\end{figure*}

To comprehensively evaluate the performance of our proposed image editing method, we conducted a user study involving 200 participants. The participant pool was recruited to represent diverse perspectives and expertise levels relevant to image manipulation tasks, comprising 60 professional graphic designers (30\%), 70 engineers with computer vision/ML experience (35\%), and 70 general users without technical backgrounds (35\%). This stratified sampling ensured balanced assessment across usability and technical robustness dimensions. Participants were presented with a randomized sequence of original and edited image pairs across varied scenarios and asked to evaluate results using a structured survey titled Image Editing Method User Evaluation Survey (as shown in \cref{fig:survey}).

The survey employed a 0-5 rating scale (0=Worst, 5=Best) across three critical dimensions: User-defined Region Flexibility assessed whether edits were confined strictly to user-specified areas (\emph{e.g., "hip-length→knee-length" transformations}), with ratings ranging from 5 for perfect localization to 0 for complete failure; Pose Robustness measured the naturalness of edits under challenging poses (\emph{e.g., extreme stances}), where 5 indicated flawless limb/joint preservation and 0 denoted severe disconnections; In-the-Wild Generalizability evaluated performance in real-world conditions (\emph{e.g., crowds, occlusions, night scenes}), with 5 representing seamless integration in cluttered backgrounds and 0 indicating catastrophic background/identity corruption. Display hardware and lighting conditions were standardized across all testing sessions to ensure consistent visual assessment.

\section{Evaluation on Wild Images}
\label{Wild Images}
\paragraph{Qualitative}
\begin{figure*}[h]
	\centering
	\begin{minipage}{1\linewidth}
		\centering
		\includegraphics[width=1\linewidth]{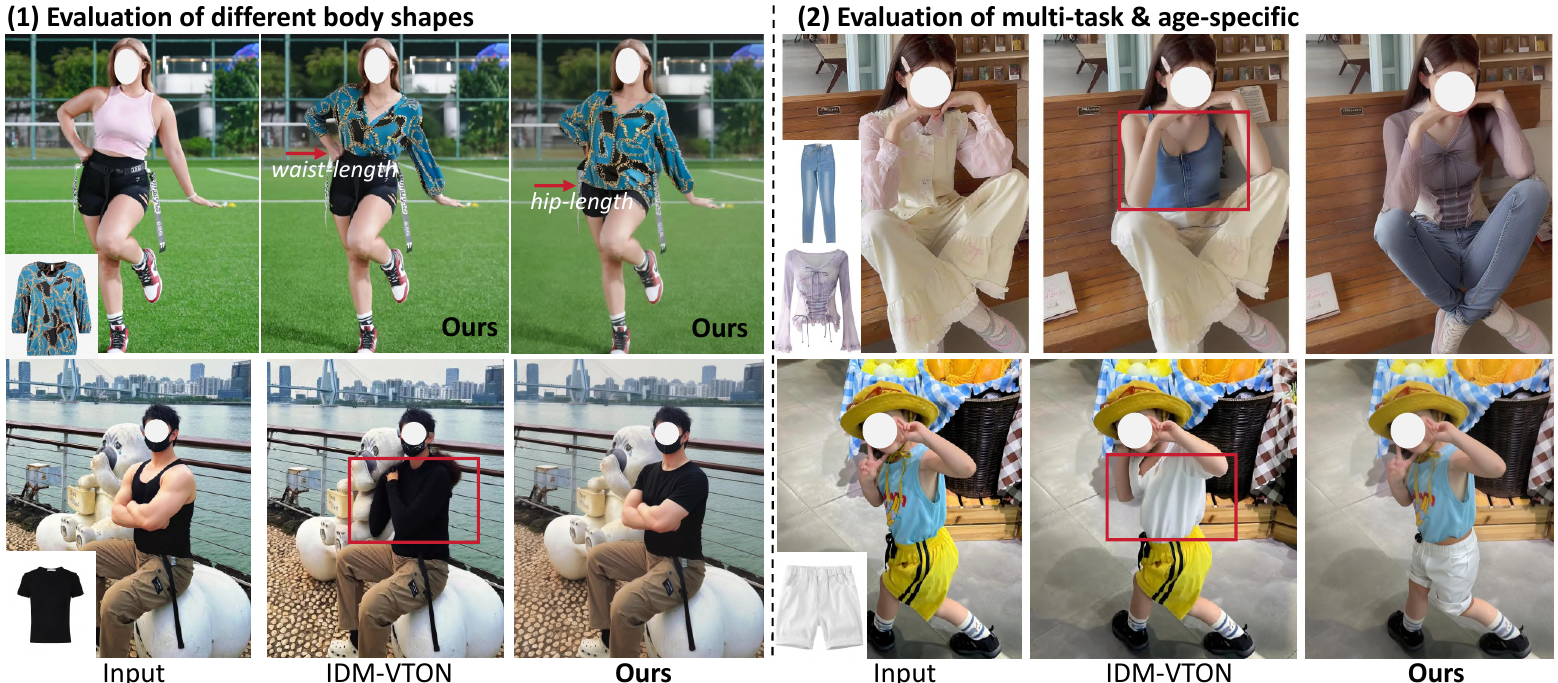}
	\end{minipage}
	
	\begin{minipage}{1\linewidth}
		\centering
		\includegraphics[width=1\linewidth]{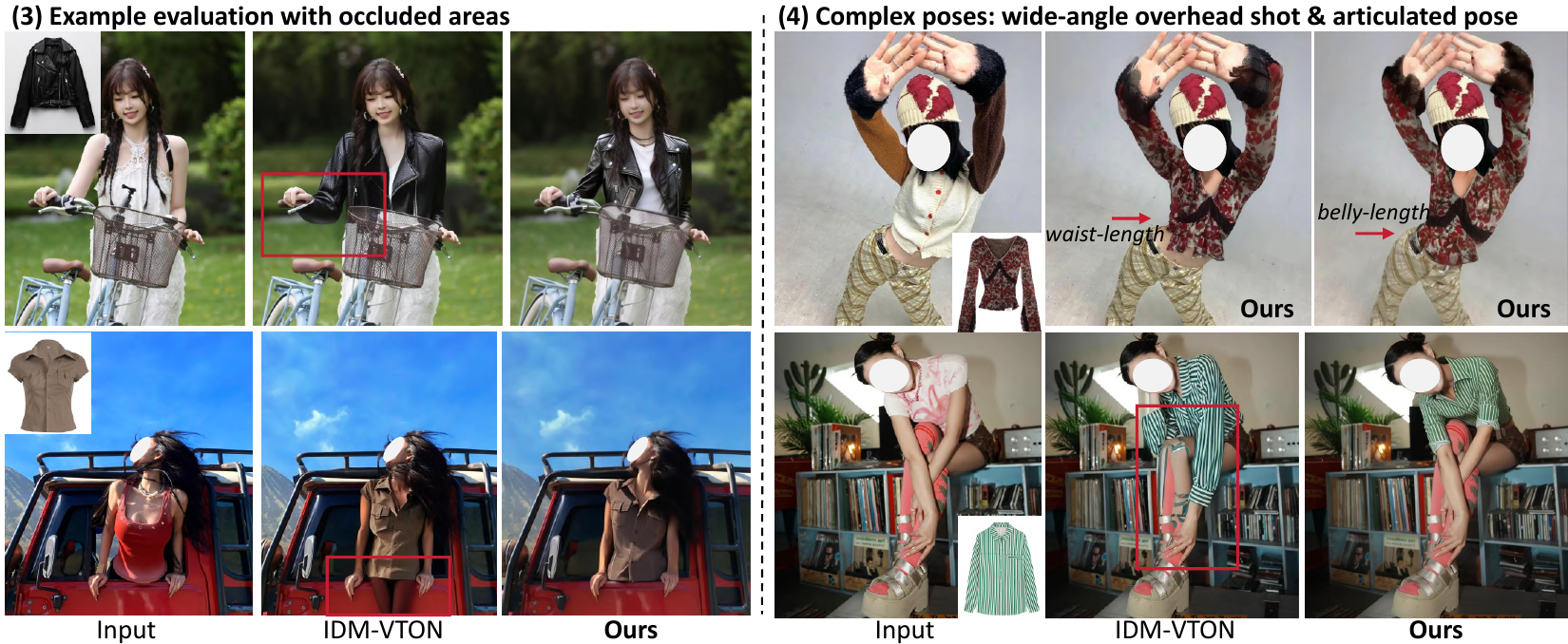}
		\caption{Multi-scenario Evaluation: Diverse Body Types, Multi-task, Age-specific, Partial Occlusion, Wide-angle Shots, Articulated Poses}
		\label{fig:scenario}
	\end{minipage}
\end{figure*}
As illustrated in \cref{fig:scenario}, we present additional evaluations across diverse scenarios: varying body types, multi-task editing, specific age groups, regional occlusions, wide-angle captures, and hinged poses. Existing methods, constrained by rigid masks, can only edit upper-body regions while failing to modify lower-body areas flexibly. Our approach not only enables precise editing of user-specified regions but also supports concurrent multi-task edits (\emph{e.g., simultaneous upper and lower garment replacement}). It faithfully preserves original body proportions across diverse physiques. In occlusion scenarios where non-editable foreground objects overlay target regions, our method retains foreground elements while plausibly editing underlying targets. Under extreme conditions—including acute hinged poses and wide-angle overhead shots—it robustly accomplishes virtual try-on tasks. These qualitative results demonstrate the robust and powerful editing capabilities of our method across diverse challenging scenarios.

\paragraph{Quantitative}
To further quantify the performance of our method, we employ ImageReward~\cite{Xu2023ImageRewardLA} as the evaluation metric. Editing results generated by different models on our collected test dataset are assessed, with mean ImageReward scores reported. As shown in \cref{tab:human}, Pose-Star achieves superior human feedback ratings compared to fixed-mask baselines, attributable to its anatomy-aware masks that faithfully adhere to user-specific editing instructions.

\begin{table*}[h]
\renewcommand{\arraystretch}{1.1}
\centering
\setlength{\tabcolsep}{2mm}{
\begin{tabular}{c|ccccc}
\toprule
\textbf{Methods} & TexFit~\cite{Wang2024TexFitTF}  & Leffa~\cite{Zhou2024LearningFF} & CatVTON~\cite{Chong2024CatVTONCI} & IDM-VTON~\cite{Choi2024ImprovingDM} & Ours \\ \hline
\textbf{ImageReward}~\cite{Xu2023ImageRewardLA} & 1.17 & 1.35  & 1.19    & 1.16     & 1.61 \\ \bottomrule
\end{tabular}
}
  \caption{
  Quantitative evaluation of human feedback. }
  \label{tab:human}
\end{table*}

\section{Discussion and Limitations}
\label{Discussion}
\begin{figure*}[h]
  \centering
   \includegraphics[width=1\linewidth]{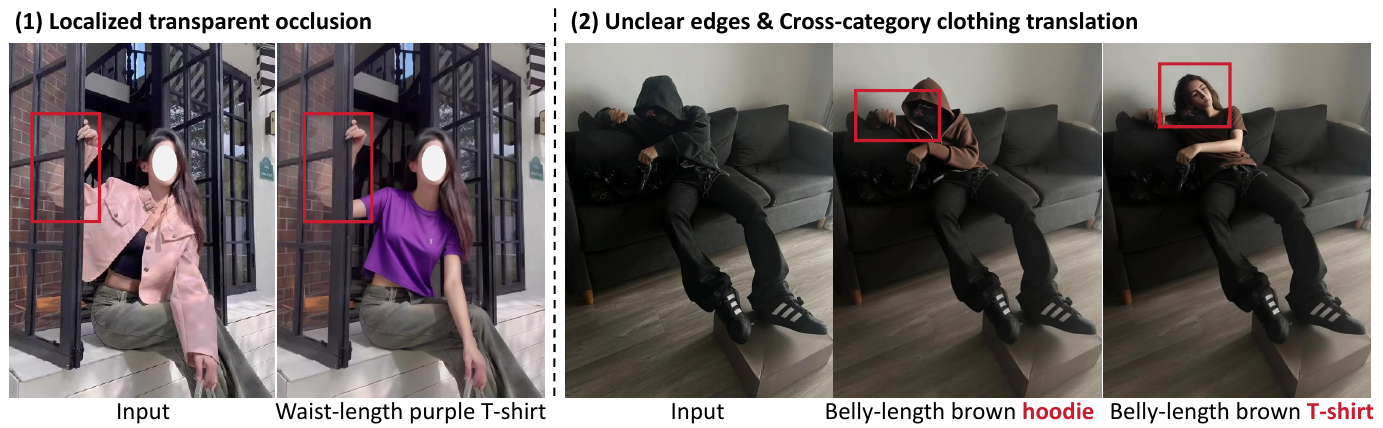}
   \caption{\textbf{ Limitations.} Foreground occlusion scene and random face ID problem.}
   \label{fig:limitations}
\end{figure*}
Pose-Star exhibits limitations on specific samples: when editing transparent occlusions (as shown in \cref{fig:limitations}), such as cases where a human arm is partially obscured by glass and the occluded area falls within the target editing region, the method struggles to recognize body parts behind the glass and fails to generate the edited target with consistent occlusion. Similar challenges arise in scenarios like wire-mesh occlusion editing or veil occlusion editing. This limitation indicates that neither Pose-Star nor existing editing models possess layer-awareness, necessitating more advanced layer-perceptive editing capabilities. Another limitation involves cross-garment category edits (\emph{e.g., "hoodie→T-shirt"}) when the original facial region is occluded: such edits cannot preserve the subject’s original facial identity, instead producing randomly generated facial identities—an undesirable outcome. This specific issue remains unaddressed in prior works; we propose that future research could incorporate additional facial identity constraints to mitigate identity loss and random generation during editing.
\end{document}